\documentclass[10pt,twocolumn,letterpaper]{article}

\usepackage[pagenumbers]{cvpr} 

\usepackage[dvipsnames]{xcolor}

\definecolor{cvprblue}{rgb}{0.21,0.49,0.74}
\usepackage[pagebackref,breaklinks,colorlinks,citecolor=cvprblue]{hyperref}

\usepackage{url}

\usepackage[utf8]{inputenc} 
\usepackage[T1]{fontenc}    
\usepackage{booktabs}       
\usepackage{amsfonts}       
\usepackage{nicefrac}       
\usepackage{microtype}      

\usepackage{physics}
\usepackage{graphicx}
\usepackage{xcolor}
\usepackage{romannum}
\usepackage{amsmath}
\usepackage{caption, subcaption}
\usepackage{epstopdf}

\usepackage{tabu}

\usepackage{wrapfig}

\newcommand{\tablestyle}[2]{\setlength{\tabcolsep}{#1}\renewcommand{\arraystretch}{#2}\centering\footnotesize}

\definecolor{baselinecolor}{gray}{.9}

\definecolor{diff_blue}{HTML}{2F5597}
\definecolor{primer_green}{HTML}{4C8485}
\definecolor{data_brown}{HTML}{C27D32}
\definecolor{approx_marginal_red}{HTML}{C00000}
\definecolor{marginal_orange}{HTML}{F4B183}

\usepackage{booktabs}
\usepackage[vlined, ruled]{algorithm2e}

\usepackage[normalem]{ulem}

\usepackage{multirow}

\def\paperID{14188} 
\def\confName{CVPR}
\def\confYear{2024}

\title{Masked Diffusion Models are Fast Distribution Learners}

\author{Jiachen Lei, Qinglong Wang, Peng Cheng, Zhongjie Ba\thanks{Corresponding author}, Zhan Qin, Zhibo Wang\\ Zhenguang Liu, Kui Ren\\
Zhejiang University, Hangzhou, China\\
{\tt\small \{jiachenlei, qinglong.wang, peng\_cheng, zhongjieba, qinzhan, zhibowang}\\
{\tt\small liuzhenguang, kuiren\}@zju.edu.cn}
}

\begin{document}
\maketitle
\begin{abstract}
Diffusion model has emerged as the \emph{de-facto} model for image generation, yet the heavy training overhead hinders its broader adoption in the research community. 
We observe that diffusion models are commonly trained to learn all fine-grained visual information from scratch. 
This paradigm may cause unnecessary training costs hence requiring in-depth investigation. 
In this work, we show that it suffices to train a strong diffusion model by first pre-training the model to learn some primer distribution that loosely characterizes the unknown real image distribution. 
Then the pre-trained model can be fine-tuned for various generation tasks efficiently. 
In the pre-training stage, we propose to mask a high proportion (e.g., up to 90\%) of input images to approximately represent the primer distribution and introduce a masked denoising score matching objective to train a model to denoise visible areas. 
In subsequent fine-tuning stage, we efficiently train diffusion model without masking. 
Utilizing the two-stage training framework, we achieves significant training acceleration and a new FID score record of 6.27 on CelebA-HQ $256 \times 256$ for ViT-based diffusion models.
The generalizability of a pre-trained model further helps building models that perform better than ones trained from scratch on different downstream datasets. 
For instance, a diffusion model pre-trained on VGGFace2 attains a 46\% quality improvement when fine-tuned on a different dataset that contains only 3000 images.
Our code is available at \url{https://github.com/jiachenlei/maskdm}.
\end{abstract}    
\section{Introduction}
\label{sec:intro}

Diffusion models~\citep{ ho2020denoising, sohl2015deep, song2020score, NCSN} have demonstrated exceptional performance in image generation and emerged as the state-of-the-art learning models for this task. 
The core denoising training approach has also been quickly adopted in various tasks such as image editing~\citep{saharia2022palette, goel2023pair, avrahami2022blended, singh2023high} and controllable image generation~\citep{dalle, dalle2, imagen, epstein2023diffusion, qiu2023controlling, ruiz2023dreambooth, chen2023training, balaji2022ediffi}. 
However, this approach is commonly adopted by training a model to simultaneously learn all fine-grained visual details presented in images throughout the entire training process, demanding intensive computational resources, especially for generating high-resolution images. 
In this work, we try to investigate if current denoising training paradigm can be improved for accelerating the overall training process by avoiding modeling complete images in the early training stage and if the improved paradigm can be applied in tandem with previous studies.

We start by describing our approach by take painting as an intuitive example. 
Rather than directly traversing all fine-grained details, a painter usually starts with more distinguishing features, such as the global structure or local prominent texture. 
We anticipate that this natural task decomposition can also be applied to train diffusion models, making training comparably easier by first approximating some "primer" distributions that preserve salient image features. 
This also leads to easier training since inspecting all intricate details usually causes much training difficulty. 
With a well pre-trained model that learns salient image features, the subsequent learning of complete detailed image information can be effectively accelerated.

However, it is non-trivial to learn such primer distributions from the real distribution which is unknown by itself. 
To address this challenge, we first define a \textit{primer} distribution as one that shares the same group of marginals, which contains diverse important features, with the target distribution, hence can be further transformed into the target distribution. 
There exist many distributions that satisfy this definition. 
In our approach, we propose a simple yet effective method to implicitly approximate this primer distribution by modeling various marginals of the target distribution. 
Specifically, we apply random masking to every image input to a diffusion model.
Each masked image can be regarded as a sample drawn from some arbitrary marginal distribution. 
We also provide to the model positional information of visible pixels as clues to distinguish different marginals. 
We consider this approach as sharing a close spirit as Dropout~\cite{srivastava2014dropout}, which essentially learns a distribution of models. 
By performing denoising training on the visible parts, we try to approximately learn a joint distribution, which is composed of various marginal distributions that can be aggregated to preserve meaningful local or global features. 

Consequently, the prevalent end-to-end process for training a diffusion model can be decomposed into a two-stage path: the first \emph{masked pre-training} stage, which enables a preferable initialization point for modeling the target distribution by performs masked denoising score matching (MDSM) on visible parts, followed by \emph{denoising fine-tuning} equipped with the conventional weighted denoising score matching (DSM) objective~\citep{ho2020denoising, DenoisingSM} in the second stage. 
The masking strategy and rate are chosen empirically as hyper-parameters and remain fixed throughout the training process. 
We name the models yielded by this training framework as \textbf{Masked Diffusion Models} (MaskDM). 
It is important to note that a sufficient masked pre-training accelerates training a model across various datasets, even when a model can only be fine-tuned with limited data. The {\bf contributions} can be summarized as follows:

(\romannum{1}) We design a two-stage training framework that integrates various masking strategies into the pre-training stage to improve the efficiency of diffusion model training. 
Through thorough experiments, we examine the effects of different masking configurations on both model performance and efficiency improvement, offering practical guidance for implementing our proposed framework. 
In particular, we apply the proposed framework to set a new record for the FID score of 6.27 on the CelebA-HQ $256 \times 256$ dataset.

(\romannum{2}) We demonstrate that masked pre-training enables reduced computational expenses when adapting pre-trained models to various datasets through fine-tuning. Specifically, we demonstrate that fine-tuning models pre-trained on different datasets yields better performance than training models from scratch on the given dataset, under the same training time and data conditions. This performance gap is even larger when the available data is limited. 

(\romannum{3}) We employ masked pre-training to significantly mitigate the training complexity of ViT-based diffusion models. Through extensive experiments, we compare the performance of models trained with and without our proposed framework. We show that the improvement in both training efficiency and model generation performance becomes increasingly evident as image resolution increases. These results help pave the way for broader adoption of the ViT architecture for constructing more powerful diffusion models.
 \section{Preliminary on diffusion models}\label{sec:background}

Training diffusion model~\citep{sohl2015deep, ho2020denoising} takes a forward and a reverse process.
The forward process is defined as a discrete Markov chain of length $T$: $q(\boldsymbol{x}_{1:T}|\boldsymbol{x}_0) = \prod_{t=1}^{T}{q(\boldsymbol{x}_t|\boldsymbol{x}_{t-1})}$.
For each step $t\in[1,T]$ in the forward process, a diffusion model adds noise $\epsilon_t$ sampled from the Gaussian distribution $\mathcal{N}(0,\mathbf{I})$ to data $\boldsymbol{x}_{t-1}$ and obtains disturbed data $\boldsymbol{x}_t$ from $q(\boldsymbol{x}_t|\boldsymbol{x}_{t-1}) = \mathcal{N}(\boldsymbol{x}_t; \sqrt{1-\beta_t} \boldsymbol{x}_{t-1}, \beta_{t}^{2}\mathbf{I}$).
$\beta$ determines the scale of added noise at each step and can be prescribed in different ways~\citep{ho2020denoising, nichol2021improved} such that $p(\boldsymbol{x}_T) \approx \mathcal{N}(0,\mathbf{I})$.
Noticeably, instead of sampling sequentially along the Markov chain, we can sample $x_t$ at any time step $t$ in the closed form via $q(\boldsymbol{x}_t | \boldsymbol{x}_0) = \mathcal{N}(\boldsymbol{x}_t;\sqrt{\bar{\alpha}_t}\boldsymbol{x}_0,(1-\bar{\alpha}_t)I)$, where $\bar{\alpha}_t=\prod_{s=1}^{t}(1-\beta_s)$. 
The reverse process is also defined as a Markov chain: $p_{\theta}(\boldsymbol{x}_{0:T})= p(\boldsymbol{x}_T)\prod_{t=1}^{T}{p_{\theta}(\boldsymbol{x}_{t-1}|\boldsymbol{x}_t)}$. In DDPM~\cite{ho2020denoising}, $p_{\theta}(\boldsymbol{x}_{t-1}|\boldsymbol{x}_{t})$ is parameterized as $\mathcal{N}(\boldsymbol{x}_t;\mu_\theta(\boldsymbol{x}_{t}, t), \sigma_t)$, where $\mu_\theta(\boldsymbol{x}_{t}, t)=\frac{1}{\sqrt{\alpha_t}}(\boldsymbol{x}_t-\frac{\beta_t}{\sqrt{1-\bar{\alpha}_t}}\boldsymbol{\epsilon}_{\theta}(\boldsymbol{x}_t, t))$ and $\sigma_t$ is a time-dependent constant. Given $\boldsymbol{x}_{t}$ and the time step $t$, $\boldsymbol{\epsilon}_{\theta}$ is a neural network and aims at predicting the noise $\boldsymbol{\epsilon} \sim \mathcal{N}(0,\mathbf{I})$ used to construct $\boldsymbol{x}_{t}$ together with $\boldsymbol{x}_{t-1}$. Using this parameterization, the variational objective in~\cite{sohl2015deep} is ultimately simplified to Eq.\ref{eq:lsimple}, which can be seen as a variant of DSM~\cite{DenoisingSM} over multiple noise scales.
\begin{equation}
    \label{eq:lsimple}
    \begin{gathered}
        L(\theta)=\mathbb{E}_{t,\boldsymbol{x}_0,\epsilon}{ \left[ \big\| \boldsymbol{\epsilon}-\boldsymbol{\epsilon_\theta}(\sqrt{\bar{\alpha}_t}\boldsymbol{x}_0+\sqrt{1-\bar{\alpha}_t}\boldsymbol{\epsilon}, t) \big\|^2 \right] }
    \end{gathered}
\end{equation}
In our two-stage training framework, we mainly adopt the above vanilla training procedure of DDPM in the fine-tuning stage. The pre-trained model can also be fine-tuned with other training framework, such as VPSDE~\cite{song2020score}, the continuous variant of DDPM (investigated in Sec.\ref{sec:generalizability}). Before the fine-tuning starts, the model is loaded with weights obtained via masked pre-training, which we introduced in Sec.\ref{sec:masked pretrain}. 

\section{Masked diffusion models}
\label{sec:mdt}

\begin{figure*}
    \begin{minipage}[t]{.73\linewidth}
        \begin{minipage}[b]{.5\linewidth}
            \centering
            \includegraphics[width=\linewidth]{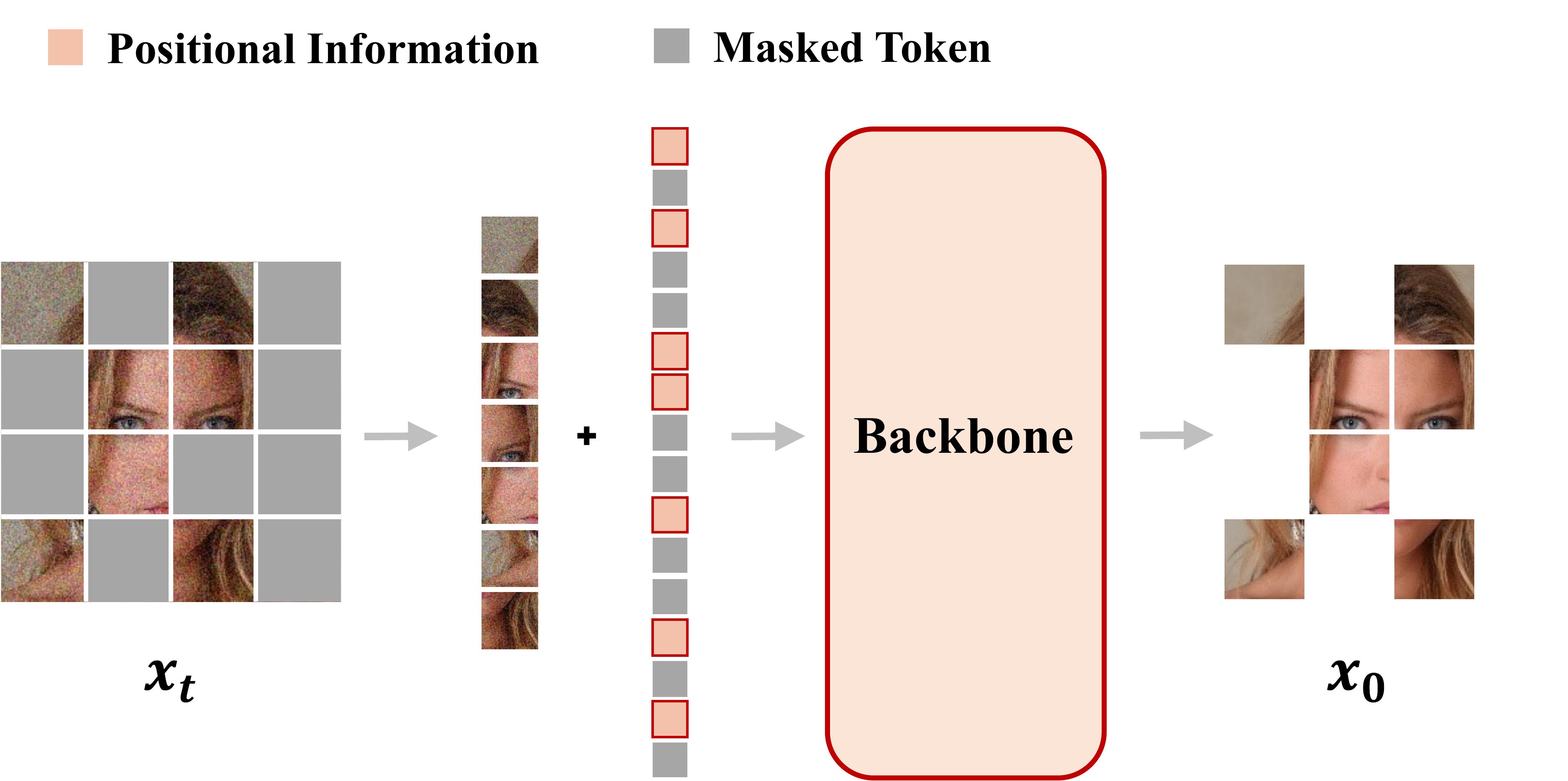}    
            \subcaption{}
        \end{minipage}%
        \begin{minipage}[b]{.5\linewidth}
            \centering
            \includegraphics[width=.9\linewidth]{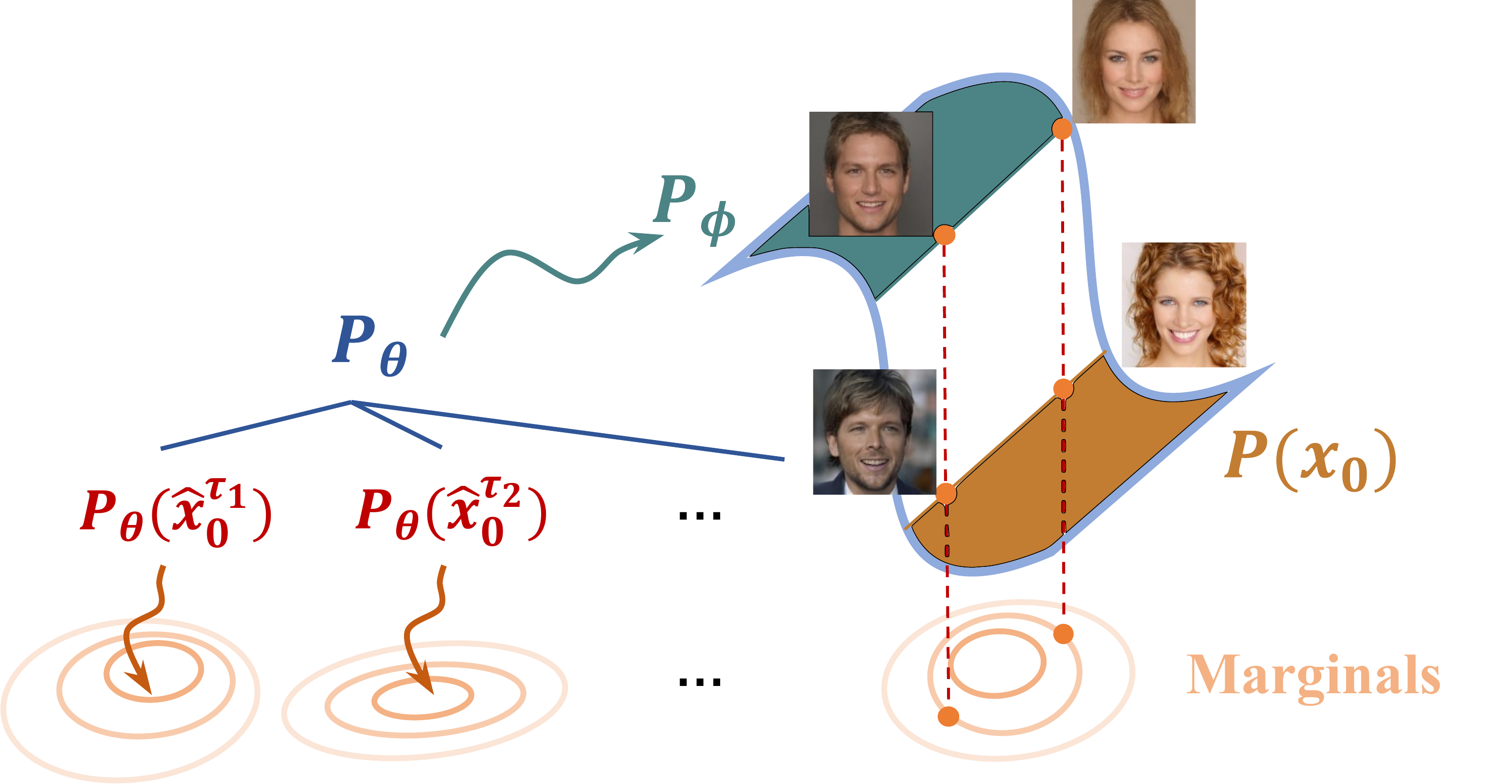}    
            \subcaption{}
        \end{minipage}%
        \caption{(a) Illustration of the pre-training process. The masked input can be viewed as a sample from {\color{marginal_orange}$p_{\phi}(\boldsymbol{x}_{0}^{\tau})$}, while {\color{approx_marginal_red} red boxes} denote the parameters of {\color{approx_marginal_red}$p_{\theta}(\boldsymbol{x}_{0}^{\tau})$}.  (b) An explanation to the process of approximating primer distribution {\color{primer_green}$p_{\phi}$}: the diffusion model {\color{diff_blue}$p_{\theta}$} approximate {\color{primer_green}$p_{\phi}$} by modeling its marginals {\color{marginal_orange}$p_{\phi}(\boldsymbol{x}_{0}^{\tau})$} via {\color{approx_marginal_red}$p_{\theta}(\boldsymbol{x}_{0}^{\tau})$}. Noticeably, {\color{primer_green}$p_{\phi}$} and the true data distribution {\color{data_brown}$p(\boldsymbol{x}_0)$} shares the same set of marginal distributions according to our definition. }
        \label{fig:architecture}
    \end{minipage}
    \hfill
    \begin{minipage}[t]{.25\linewidth}
        \centering
        \includegraphics[width=.9\linewidth]{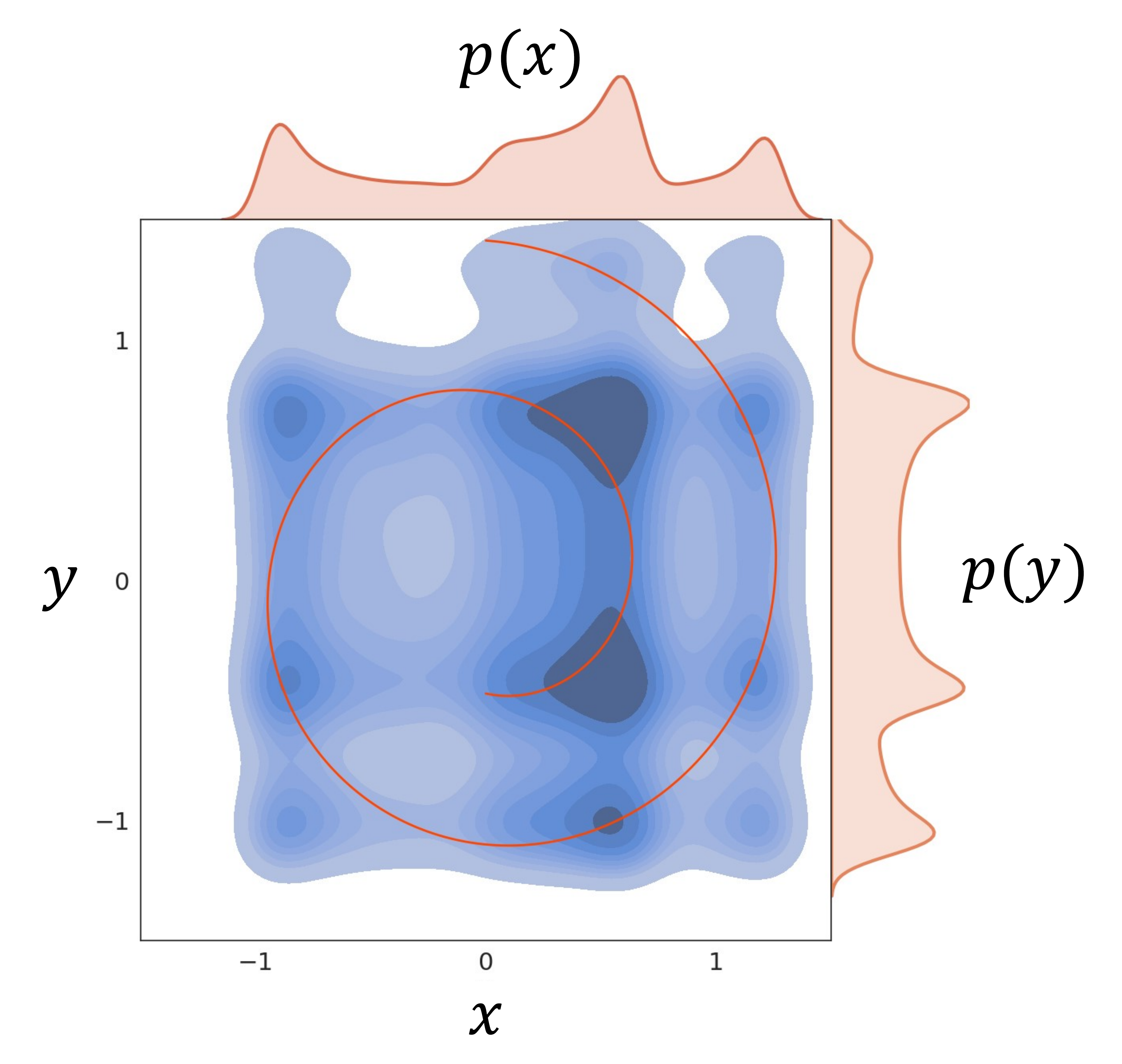}
        \captionof{figure}{2D Swiss roll example. The red spiral line denotes the true data distribution while the blue heatmap is a learned primer distribution.}
        \label{fig:toyexample}
    \end{minipage}
\end{figure*}

We present an intuitive explanation for our design in Fig.~\ref{fig:toyexample}. 
Assuming that we are approximating a 2D Swiss roll distribution $p(\boldsymbol{z})$  (marked by the red spiral line), where $\boldsymbol{z}=(x, y)$. There is another distribution $p_{\phi}(\boldsymbol{z})$ (represented as a blue heatmap) that fully covers $p(\boldsymbol{z})$ by traversing all modes. 
Rather than directly approximating $p(\boldsymbol{z})$, it is expected to be comparably easier to gradually transform an initial distribution $p_{\phi}(\boldsymbol{z})$, which shares with $p(\boldsymbol{z})$ the same marginal distribution, i.e., $p(x)$ and $p(y)$, into 
$p(\boldsymbol{z})$. 
For image data, as the dimensionality increases, the data space expands significantly faster than the space expanded by real image samples. 
As such, approximating a high-dimensional $p(\boldsymbol{z})$ with $p_{\phi}(\boldsymbol{z})$, which partially preserves the sophisticated relations between different marginal distributions, may bring even more computational benefits.

\subsection{Masked pre-training}
\label{sec:masked pretrain}
We denote an image $\boldsymbol{x_0}$\footnote{We follow the conventions and denote a clean image as $\boldsymbol{x_0}$, where the subscript $0$ is the time step.} by a vector: $(x^1_0, x^2_0,, x^3_0,, ... , x^N_0)$ , where $N$ represents the number of pixels. Then the data distribution $p(\boldsymbol{x_0})$ can be expressed as the joint distribution of $N$ pixels. Let $\tau$ represents a randomly selected subsequence of $[1, ..., N]$ with a length of $S$. We denote the subset of selected pixels as $\{x_0^{\tau_i}\}^S_{i=1}$ and the resulting marginal distribution as $p(\boldsymbol{\hat{x}_0^\tau}) = p(x_0^{\tau_1}, x_0^{\tau_2}, x_0^{\tau_3}, ..., x_0^{\tau_S})$. For simplicity, with $S$ being fixed, we utilize $\boldsymbol{\hat{x}_0}$ to represent any marginal variable combinations $\{ \tau \in [1, ..., N], |\tau|=S\ |\ \boldsymbol{\hat{x}_0}^\tau \}$, and $p(\boldsymbol{\hat{x}_0})$ to represent the corresponding marginal distribution. It is evident that $p(\boldsymbol{x_0})$ belongs to a family $\mathcal{Q}$ of distributions that share the same set of marginals $p(\boldsymbol{\hat{x}_0})$. We introduce the term \textit{primer} distribution to refer to any distribution in $\mathcal{Q}$ other than $p(\boldsymbol{x_0})$ that satisfies this condition. We represent such distributions using the notation $p_{\phi}(\boldsymbol{x_0})$, where $\phi$ represents the unknown true distribution parameters.

It is non-trivial to approximate $p_{\phi}(\boldsymbol{x_0})$, particularly when the samples from $p_{\phi}(\boldsymbol{x_0})$ are not available. We initialize the task of approximating $p_{\phi}(\boldsymbol{x_0})$ with a diffusion model $p_\theta(\boldsymbol{x_0})$, defined as introduced in Sec.\ref{sec:background}. In each training iteration, by training with a batch of images sampled from some arbitrary marginal distributions, which can be further viewed as sampled from $p_\theta(\boldsymbol{x_0})$, we are implicitly approximating $p_{\phi}(\boldsymbol{x_0})$ by modeling its various marginals. 

To achieve this, we mask each image input $\boldsymbol{x_0}$ with a vector $\mathbf{M}\in \{0,1\}^{N}$, and incorporate the positional information $\mathbf{H}\in R^{N}$ of the visible pixels into the model input as additional clues to distinguish different marginal distributions.
Therefore, the model input becomes $\boldsymbol{\hat{x}_0}=\textbf{M} \odot (\boldsymbol{x_0} + \mathbf{H})$ and the noise is $\boldsymbol{\hat{\epsilon}}=\textbf{M} \odot (\boldsymbol{\epsilon} + \mathbf{H})$.
In practice, we observe that this simple masking approach suffices to preserve meaningful visual details while enabling a much faster pre-training convergence.
Furthermore, it facilitates the subsequent fine-tuning, hence reducing the overall training time.
The masked image $\boldsymbol{\hat{x}_0}$ and noise $\boldsymbol{\hat{\epsilon}}$ are then integrated to construct $\boldsymbol{\hat{x}_t}$ such that $\boldsymbol{\hat{x}_t}=\sqrt{\bar{\alpha}_t}\boldsymbol{\hat{x}_0}+\sqrt{1-\bar{\alpha}_t}\boldsymbol{\hat{\epsilon}}$. Then we substitute $x_0$ and $\epsilon$ defined in Eq.~\ref{eq:lsimple} with $\hat{x}_0$ and $\hat{\epsilon}$, respectively, to optimize model parameters. For notation clarity, we name the updated objective as masked denoising score matching (MDSM), as presented in Eq.~\ref{msm_lsimple}:
\begin{equation}
    \label{msm_lsimple}
    \begin{gathered}
         L(\theta)=\mathbb{E}_{t,\boldsymbol{\hat{x}_0},\boldsymbol{\hat{\epsilon}}}{ \left[ \big\| \boldsymbol{\hat{\epsilon}}-\boldsymbol{\epsilon_\theta}(\sqrt{\bar{\alpha}_t}\boldsymbol{\hat{x}_0}+\sqrt{1-\bar{\alpha}_t}\boldsymbol{\hat{\epsilon}}, t) \big\|^2 \right] }
    \end{gathered}
\end{equation}

In the working pipeline overview, as illustrated in Fig.\ref{fig:architecture}, we present an example use case where a face image is masked with a set of grey patches. 
The masked image can be seen as a sample drawn from a marginal distribution that is identified by the selected square blocks, which  marginalize out all covered pixels. 
Considering the positional information $H$ as some fixed or learnable parameters of the model, then $p_{\theta}(\boldsymbol{x_0})$ is also "marginalized" by applying masking to subsample $H$.
As such, given sufficient training time, $p_\theta(\boldsymbol{x_0})$ converges to a certain primer distribution $p_{\phi}(\boldsymbol{x_0})$ from $\mathcal{Q}$, based on which we further approximate the true data distribution $p(\boldsymbol{x_0})$ via fine-grained denoising training. The details are discussed in Sec.\ref{sec:background}.
Additionally, following the conventional sampling procedure~\citep{ho2020denoising, song2020denoising} of diffusion models, we could draw samples from $p_{\theta}(\boldsymbol{x_0})$ or its marginal distributions by customizing $\mathbf{M}$.

\subsection{Model architecture and masking configuration}\label{sec:msm}

Recent studies have grown interest in adopting ViT in building diffusion models. 
This is much due to facts that this architecture allows easy scaling up~\cite{peebles2022scalable} and is compatible with different data modalities~\cite{bao2023one}. 
Note that there are different ViT variants, e.g., U-ViT~\cite{bao2022all} and DiT~\cite{peebles2022scalable}. 
In our implementation, we choose U-ViT as the backbone considering its simpler architecture design~\footnote{In our early experiments, we observe that DiT and U-ViT achieves similar results.}. 
Nevertheless, the substantial computational burden, including high CUDA memory usage and lengthy training times, imposes a predominant challenge in training ViT-based diffusion models.
In our experiments, we find that our masked pre-training significantly boosts the training efficiency of ViT-based diffusion models (Fig.~\ref{fig:celebahq_comparison}).

In practice, it is crucial to carefully configure the masking configuration, including both $S$ (or the mask rate $m=1-\frac{S}{N}$) and the strategy for sampling the mask vector $\mathbf{M}$. Specifically, the mask rate $m$ determines the average degree of similarity between the true data distribution and the primer distributions such that a lower value of $m$ indicates a greater resemblance.
Besides, given U-ViT as the backbone, a mask is sampled as a group of neighbouring pixels instead of individual and independent pixels. 
As such, the sampled masks essentially determine the range of primer distributions that could be possibly learned. As illustrations, Fig.\ref{fig:mask_type_patch} and Fig.\ref{fig:mask_type_block} display various samples from two different primer distributions, which are implicitly learned via different mask sampling strategies. 

In this work, we have designed three different masking strategies, namely, patch-wise masking, block-wise masking, and cropping. Examples for each masking type are shown in Fig.~\ref{fig:mask_type}.
Patch-wise masking entails the random occlusion of a predefined number of image patches. 
Block-wise masking involves randomly selecting image blocks for masking, where each block comprises a fixed quantity of image patches. 
Lastly, cropping entails randomly selecting a top-left coordinate and the corresponding fixed-size square region then masking the area outside the chosen square. 
We explore and compare a range of configurations in Sec.\ref{sec:ablation}

\section{Experiments}

\subsection{Experimental setup}

\textbf{Implementation details.}
We compare with existing methods on three datasets: CelebA~\cite{celeba}, LSUN Church~\cite{yu2015lsun} and CelebA-HQ~\cite{karras2017progressive}.
We resize images from CelebA to $64 \times 64$ and $128 \times 128$, images from LSUN Church to $64 \times 64$, and images from CelebA-HQ to $256 \times 256$.
We implemented MaskDM models with the U-ViT architecture introduced in~\cite{bao2022all} with certain modifications.
Specifically, we utilize the U-ViT-Small setup from~\cite{bao2022all} to build MaskDM-S models and remove five transformer blocks from U-ViT-Mid to build MaskDM-B models. 
This allows a MaskDM-B model can be fitted in 1 Tesla V100 GPU card given a single $256\times 256$ image as input with a $4\times 4$ patch size. 
In all MaskDMs, we discard the appending convolutional blocks initially appearing in the U-ViT model and find the performance to be trivially affected.
Unless specified, we adopt $2\times 2$ block-wise masking when pre-training on $64\times 64$ images, and $4\times 4$ block-wise masking on images with resolution equal to or greater than $128\times 128$. Further detailed information is provided in the Appendix~\ref{sec:implementation_details}.

\textbf{Evaluation settings.} 
During the evaluation, we follow the convention and utilize Fréchet Inception Distance(FID)~\cite{heusel2017gans} to measure the quality of generated images.
We mainly employ two different samplers, namely, Euler-Maruyama SDE sampler~\cite{song2020score} and DDIM~\cite{song2020denoising}, to generate samples. When comparing with current methods, we compute FID scores on 50k generated samples, and we apply Euler-Maruyama SDE sampler with 1k sampling steps on CelebA $64 \times 64$ and DDIM sampler with 500 sampling steps on LSUN Church $64\times 64$, CelebA $128 \times 128$, and CelebA-HQ $256 \times 256$. 

\begin{figure*}[h!]

    \begin{minipage}[t]{\linewidth}
        \begin{subfigure}[t]{.34\linewidth}
            \centering
            \includegraphics[width=.98\linewidth]{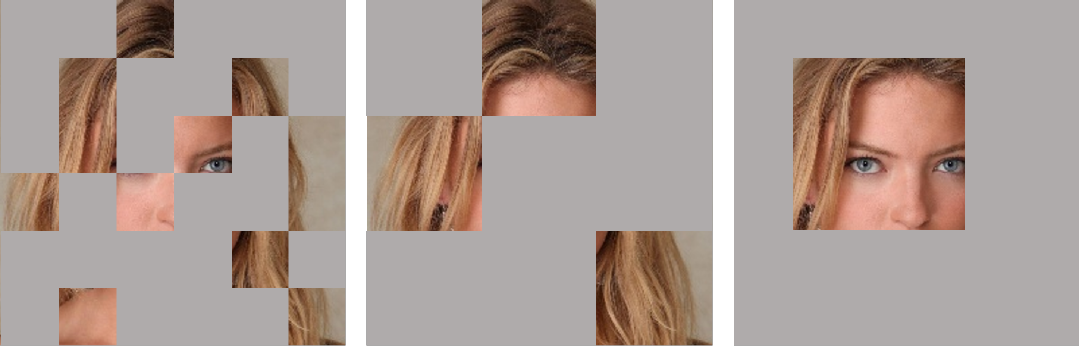}
            \caption{}
            \label{fig:mask_type}
        \end{subfigure}%
        \begin{subfigure}[t]{.32\linewidth}
            \centering
            \includegraphics[width=\linewidth]{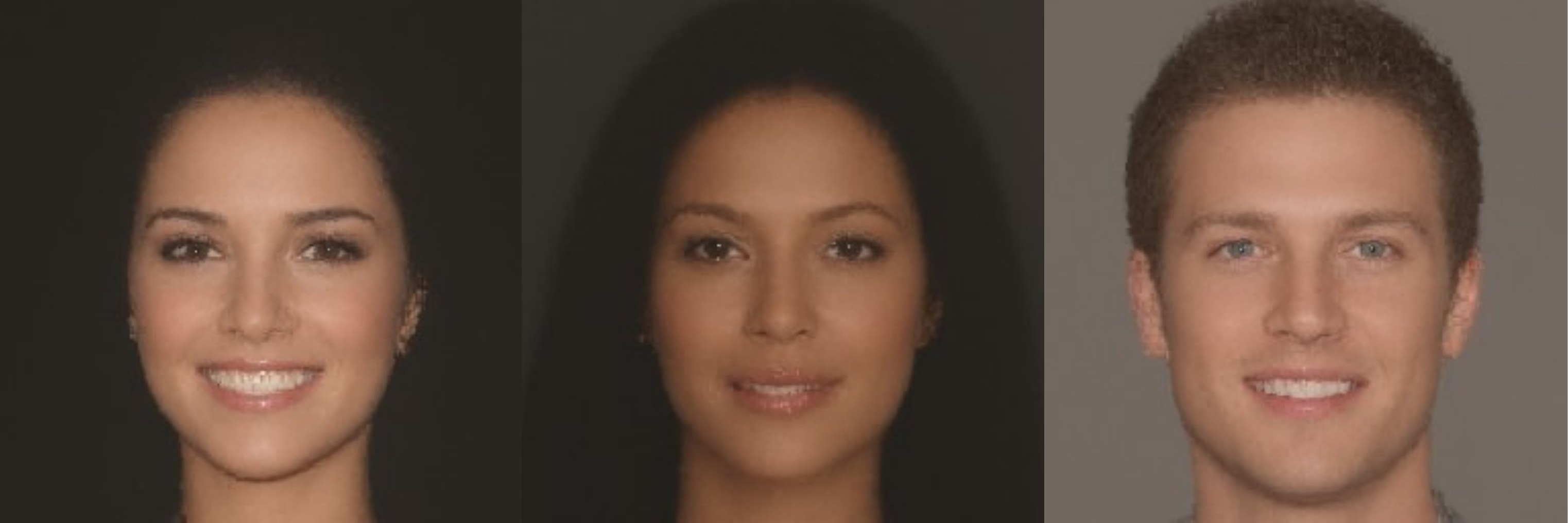}
            \caption{}
            \label{fig:mask_type_patch}
        \end{subfigure} %
        \begin{subfigure}[t]{.32\linewidth}
            \centering
            \includegraphics[width=\linewidth]{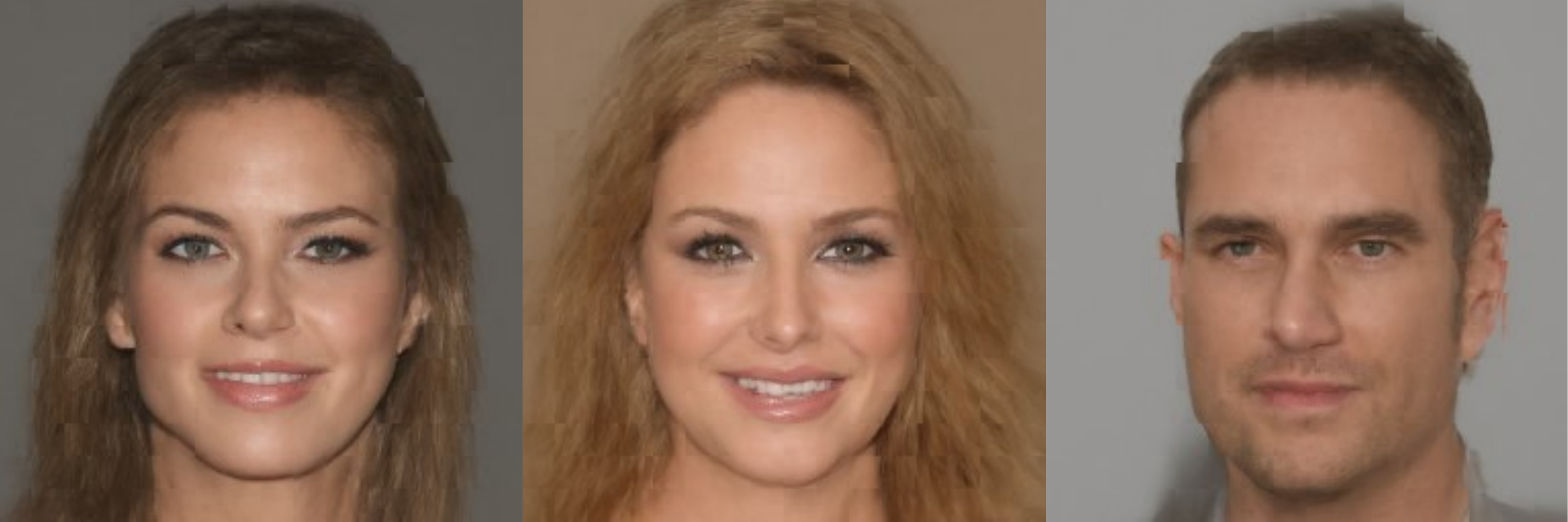}
            \caption{}
            \label{fig:mask_type_block}
        \end{subfigure}
    \end{minipage}
    \caption{(a) Three masking strategies are examined in our experiments. From left to right, the strategies are represented as patch-wise masking, block-wise masking, and cropping. (b) and (c) Samples from primer distribution, captured utilizing patch-wise and block-wise masking respectively, given a mask rate of 90\%. Notably, the model pre-trained with cropping at 90\% mask rate exhibits limited capability in generating plausible samples; therefore, we do not illustrate the results here (Please refer to the Appendix\ref{fig:appdix_cropping}).}
    \label{fig:mask_type_all}
\end{figure*}

\subsection{Investigating mask configurations}\label{sec:ablation}

\begin{table*}[h!]
    \small
    \caption{Mask configuration investigation on CelebA $64\times64$, where pre-trained weights are acquired from different masking configurations.
    The baseline model, trained from scratch without loading pre-trained weights, is marked in \colorbox{baselinecolor}{gray}. }
    \label{tab:ablation}

    \subfloat[
        Impact of mask configuration
    ]{
        \begin{minipage}[b]{0.33\linewidth}
            \centering
            \label{tab:ablation_1}
            \begin{tabular}{c | c c c }
                Mask      & 10\%  & 50\% & 90\% \\
                \toprule
                patch     & 6.85 & 6.58 & 7.34 \\
                2x2 block & \textbf{6.77} & \textbf{6.51} & 8.99 \\
                4x4 block & 6.92 & 6.88 & \textbf{6.91} \\
                cropping  & 6.92 & 6.82 & 8.62\\
                from scratch & \multicolumn{2}{c}{\colorbox{baselinecolor}{7.55}} \\

            \end{tabular}
        \end{minipage}
    }\subfloat[
       Impact of computational budget
    ]{
        \begin{minipage}[b]{0.40\linewidth}
            \centering  
            \label{tab:ablation_2}
            \begin{tabular}{c c c c c}
                Mask & Rate & Steps & bs=128 & bs=256 \\
                \toprule
                 patch     & 10\% & 50k & 6.85 & 6.31 \\
                 2x2 block & 10\% & 50k & 6.77 & 6.71 \\
                 2x2 block & 50\% & 50k & - & 6.51 \\ 
                 2x2 block & 50\% & 100k & - & 6.27 \\ 
                 2x2 block & 50\% & 150k & - & 6.05 \\
    
            \end{tabular}
            \label{tab:overhead}
        \end{minipage}
    }\subfloat[
       Impact of block size
    ]{
        \begin{minipage}[b]{0.25\linewidth}
            \centering  
            \label{tab:ablation_3}
            \begin{tabular}{c c}

                 Mask & FID$\downarrow$ \\
                 \toprule

                 patch & 6.58 \\
                 2x2 block & 6.51 \\ 
                 4x4 block & 6.88 \\ 
                 8x8 block & 7.43 \\ 
   
    
            \end{tabular}
            \label{tab:overhead}
        \end{minipage}
    }
\vspace{-.5em}
\end{table*}

To investigate the impact of different masking strategies, we adjust masking granularity and experiment with patch-wise masking, block-wise masking, and cropping, as demonstrated in Fig.~\ref{fig:mask_type}. 
Specifically, we pre-train models with different configurations on CelebA $64\times64$, using mask rates of 10\%, 50\%, and 90\%, respectively.
During pre-training, the GPU memory usage is fixed across different experiments and the default pre-training iterations are set as 50k in all experiments, 
for which we observe the pre-training curves are saturated. 
Subsequently, given a pre-trained model, we fine-tune it for 200k steps to optimize the objective delineated in Eq.~\ref{eq:lsimple}. To demonstrate the effect of adopting masked pre-training, we setup a baseline model which is trained on the same set of data from scratch (as detailed in Sec.~\ref{sec:background}) for 250k steps. 
This ensures comparable total training costs between the baseline model and its pre-trained counterpart.

\textbf{Comparing different masking types and mask rates.} 
As shown in Tab.~\ref{tab:ablation}(a), we first observe that models trained with cropping generally obtain the worst FID scores as we vary the mask rate.
A possible explanation is that randomly cropped images retain limited global structural information, which constraints the model from building long-range connections among different variables. 
As a result, given a pair of fixed batch size and training step, cropping makes it more challenging for a diffusion model to capture the consistent critical visual features. 
On the other hand, block-wise masking (including both 2x2 and 4x4 block) achieves the best results across all settings, while patch-wise masking achieves the second best FID scores.

By comparing the FID scores obtained by selecting different mask rates, we observe that pre-trained models paired with the 50\% mask rate outperform pre-trained models that adopt other masking rates in most cases. 
In particular, the model pre-trained with 2x2 block-wise masking and a 50\% masking rate achieves an FID score of 6.51, which is significantly better than the baseline. 
We also notice that models pre-trained with a 90\% mask rate exhibit a rapid divergence in FID scores after 50k training steps. 
We delve deeper into the causes of this problem by studying different influential factors (detailed findings are presented in Appendix~\ref{section:appdix_stability}.)
and find that the adopted linear noise schedule contributes significantly to training instability. This can be effectively mitigated by utilizing the cosine noise schedule~\citep{nichol2021improved}.

Moreover, the results presented in Tab.~\ref{tab:ablation}(a) show that different block sizes also impact the generation performance. 
By taking the patch-wise masking as a 1x1 block-wise masking, we compare different block sizes and demonstrate the results in Tab.~\ref{tab:ablation}(c). 
We observe that using mask with a larger block size generally leads to performance degradation. 
Therefore, in the following, we mainly focus on evaluating patch-wise and block-wise masking approaches with different mask rates.

\textbf{Delving into mask rate, batch size, and training steps.} 
In our experiments, we are particularly interested in the case where the overall computation resource is limited, which is common in academic research. 
More concretely, we assume there is a fixed GPU resource budget. 
Given this resource constraint, we confront a trade-off between mask rate and batch size. 
For instance, to maintain a constant GPU usage, when applying a lower mask rate, which consumes more CUDA memory per image, we are limited to pre-train a model with a smaller batch of data. 
Indeed, in our experiments, models using a mask rate of 10\% consume 1.5$\times$ more GPUs than those using a mask rate of 50\%. 
This raises the question that the less competitive performance obtained by block-wise and patch-wise masking with a rate of 10\% may result from an inadequate batch size of 128. 
To investigate this question, we enlarge the batch size for both aforementioned settings to 256. 
This setting corresponds to the case where the computation resources are sufficient to support larger batch size pre-training. 
As presented in Tab.~\ref{tab:ablation}(b), both models trained with block-wise and patch-wise masking with a rate of 10\% and a batch size of 256 exhibit improved performance as expected. 

We also find that a higher mask rate often requires fewer computing resources (i.e., GPUs) but slightly more training steps to achieve performance comparable to its lower mask rate counterparts. 
As such, we return to the case where the GPU memory capacity constraint still holds and confine to the previous best setting with a 2x2 block-wise masking and a mask rate of 50\%. 
We continue upon the above investigation to employing a batch size of 256, exploring the effect of extending the resource constraint in terms of pre-training steps. 
Specifically, we increase the number of pre-training steps from 50k to 100k and 150k, respectively, and present the results in Tab.~\ref{tab:ablation}(b). 
We observe a clear trend of performance improvement as the number of pre-training steps increase. 
The results are in alignment with our expectation and indicate that a longer pre-training time is generally helpful for improving the overall training performance. 

The above investigation indicates the importance of properly configuring the mask rate, batch size, and training step, for optimizing model performance while aligning with affordable computing resources. 
These empirical findings open the opportunity for designing an automated dynamic training schedule, similar to Successive Halving~\cite{jamieson2016non}, that balances the trade-off between these intertwined hyper-parameters under a constant training budget. 
In fact, we have explored manually adjusting the training schedule and obtained the best generation performance in Sec.~\ref{sec:unconditional}.  
We leave a more systematic study of training schedule automation to our future work.

\begin{table*}
    \begin{minipage}[t]{.4\linewidth}
        \vspace{0pt}
        \begin{minipage}[t]{\linewidth}
            \centering
            \captionof{table}{FID results on CelebA. Expense is measured in V100 days.$\dagger$: converted into V100 days. $\ddagger$: estimated in V100 days}
            \label{tab:celeba_comparison}
            \label{tab:celebahq128}
            \begin{tabular}{l c c c}
                Method  & FID $\downarrow$ & Params & Expense\\
                \midrule
                \textit{$64\times64$} \\
                \midrule
                DDIM~\cite{song2020denoising}   & 3.26  & 79M & - \\
                PNDM~\cite{liu2022pseudo}   & 2.71  & 79M & - \\
                U-ViT~\cite{bao2022all}  & 2.87  & 44M  & 2.08$^\ddagger$ \\
                \textbf{MaskDM-S}  & \textbf{2.27}  & 44M & 2.09\\
                
                \midrule
                \textit{$128\times128$} \\
                \midrule
                Gen-ViT~\cite{genvit}   & 22.07 & 12.9M & -\\   
                U-ViT & 12.96  & 102M & 11.67\\
                \textbf{MaskDM-B}  & \textbf{6.83}  & 102M & 8.7$^\dagger$  \\

                \bottomrule
                \\
            \end{tabular}
        \end{minipage}
        \begin{minipage}[t]{\linewidth}
            \centering
            \captionof{table}{FID results on LSUN $64\times64$. Expense is measured in V100 days}
            \label{tab:lsun_comparison}
            \begin{tabular}{l c c c}
                Method  & FID $\downarrow$ & Params & Expense\\
                \toprule  
                U-ViT & 6.58  & 44M & 1.87 \\
                \textbf{MaskDM-S}  & \textbf{5.04}  & 44M & 1.80 \\

                \bottomrule
                \\
            \end{tabular}
        \end{minipage}
    \end{minipage}
    \hfill
    \begin{minipage}[t]{.55\linewidth}
        \centering
        \vspace{0pt}
        \captionof{table}{FID results on CelebA-HQ $256\times 256$.
        Results of latent diffusion models are listed in {\color{gray}gray} color. Expense is measured in A100 days. $\ddagger$: estimated in V100 days. $\ast$: utilize extra model.
        }
        \label{tab:celebahq_comparison}
        \begin{tabular}{l c c c}
            Method  & FID $\downarrow$ & Params & Expense \\
            \toprule
            VQ-GAN~\cite{esser2021taming} & 10.2 & 355M & - \\
            PGGAN~\cite{karras2017progressive} & 8.03 & - & -\\
            DDGAN~\cite{xiao2021tackling} & 7.64 & - & -\\
            \midrule
            \textit{Latent diffusion models} \\
            \midrule
            LSGM$^\ast$~\cite{vahdat2021score} & {\color{gray}7.22} & - & -\\
            LDM-4$^\ast$~\cite{rombach2022high} & {\color{gray}5.11} & {\color{gray}274M} & {\color{gray}14.4$^\ddagger$}\\
            \midrule
            \textit{CNN-based diffusion models} \\
            \midrule
            VESDE~\cite{song2020score} & 7.23 & 66M & -\\
            Soft Truncation~\cite{kim2021soft}  & 7.16 & 66M & - \\
            P2 Weighting~\cite{choi2022perception} & 6.91 & 94M & 18.75$^\ddagger$ \\
            \midrule
            \textit{ViT-based diffusion models} \\
            \midrule
            U-ViT & 24.83 & 102M & 18.28\\
            \textbf{MaskDM-B} & \textbf{6.27} & 102M & 12.19 \\
            \bottomrule 
        \end{tabular}
    \end{minipage}
\end{table*}

\begin{figure*}
    \centering
    \includegraphics[width=\linewidth]{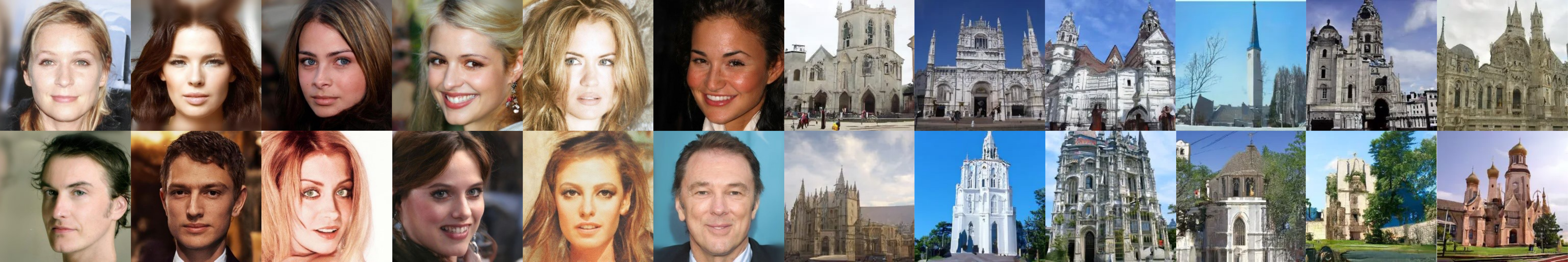}
    \caption{Qualitative results on CelebA-HQ $256\times 256$ and LSUN Church $256\times 256$}
    \label{fig:enter-label}
\end{figure*}

\begin{figure*}
    \centering
    \begin{minipage}{\linewidth}
        
            \begin{subfigure}{.24\linewidth}
                \includegraphics[width=\linewidth]{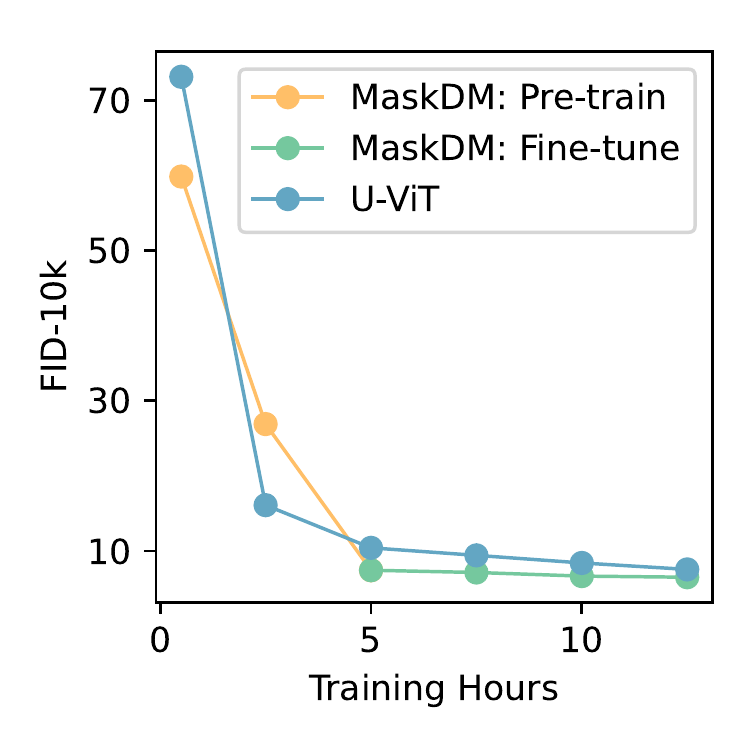}
                \vspace{-1.5em}
                \caption{CelebA $64 \times 64$}
            \end{subfigure}
            \begin{subfigure}{.24\linewidth}
                \includegraphics[width=\linewidth]{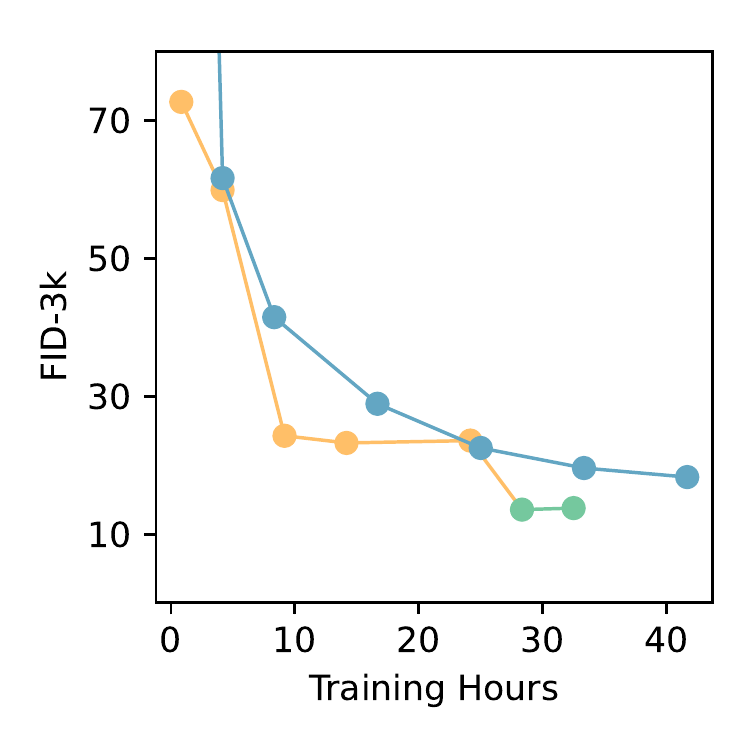}
                 \vspace{-1.5em}
               \caption{CelebA $128 \times 128$}
            \end{subfigure}
             \begin{subfigure}{.24\linewidth}
                \includegraphics[width=\linewidth]{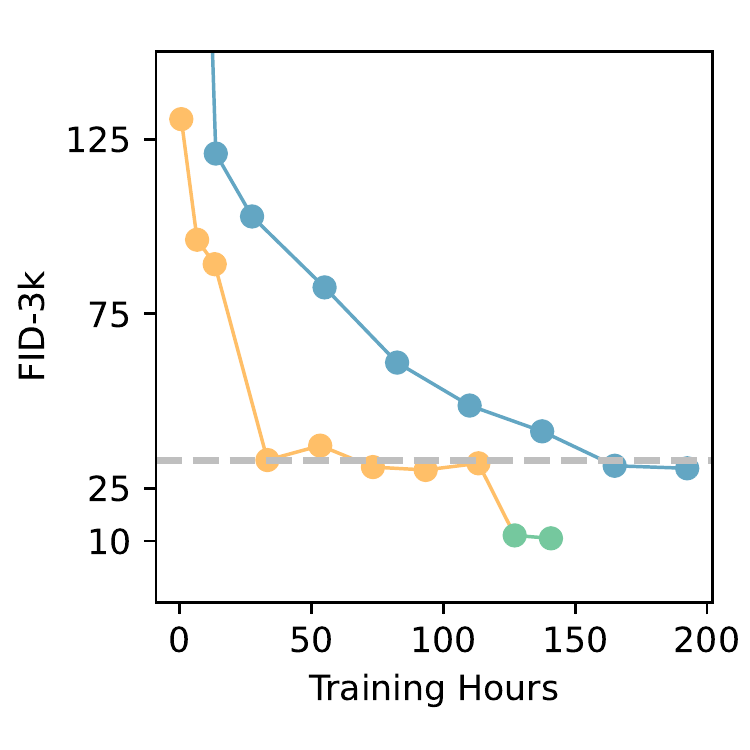}
                 \vspace{-1.5em}
                \caption{CelebA-HQ $256 \times 256$}
            \end{subfigure}%
             \begin{subfigure}{.24\linewidth}
                \includegraphics[width=\linewidth]{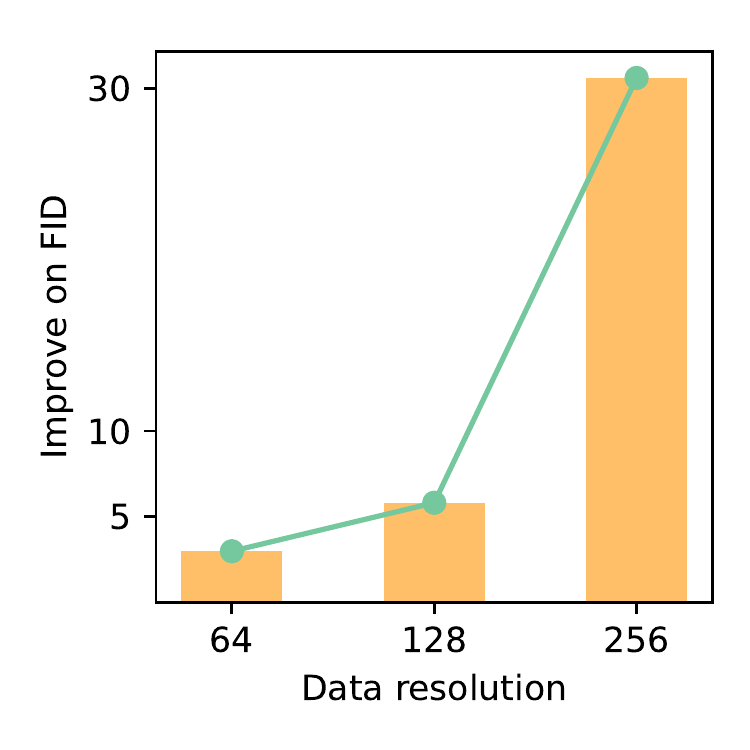}
                 \vspace{-1.5em}
                \caption{The improvement on FID}
            \end{subfigure}%
            \caption{Comparison of training efficiency between the baseline model and MaskDM on data of varying resolutions. Noticeably, the reduction in training time increases as the resolution of data increases.}
            \label{fig:celebahq_comparison}
        
    \end{minipage}
\end{figure*}

\begin{figure*}
     \begin{minipage}[t]{.5\linewidth}
         \vspace{0pt}
        \includegraphics[width=\linewidth]{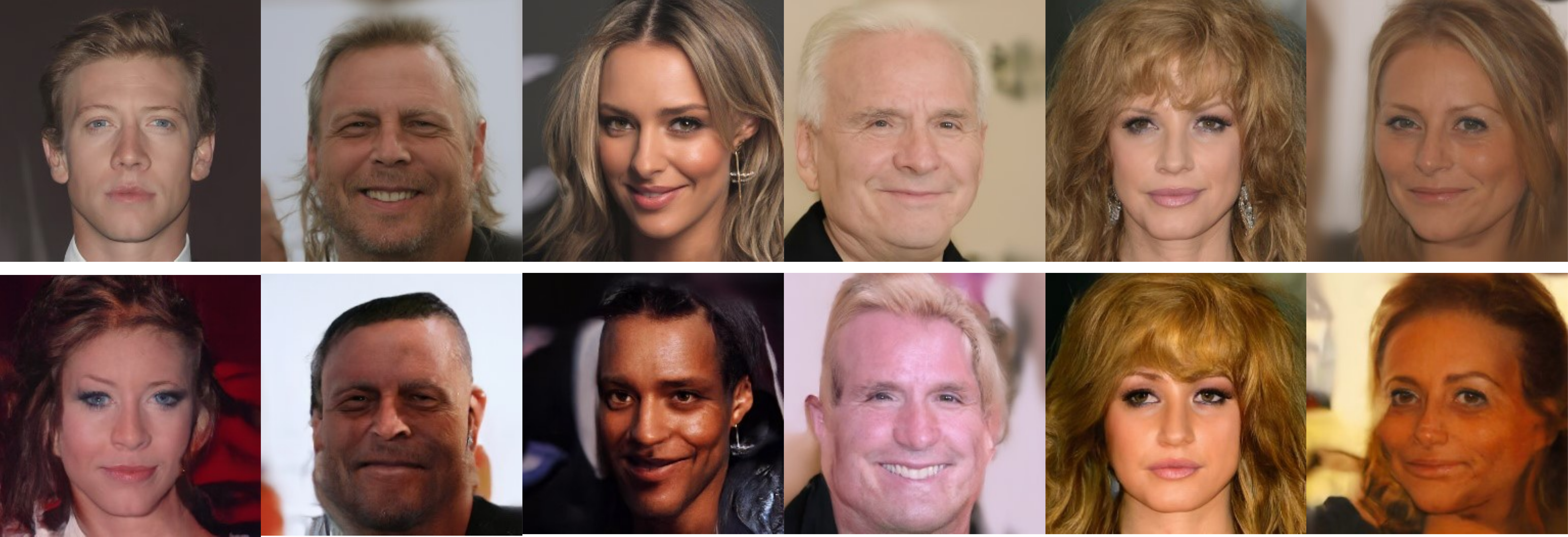}
        \caption{Qualitative comparison between our pre-trained model (top row) and the U-ViT baseline (bottom row) for reaching the same FID score, as indicated by the gray line in Fig.\ref{fig:celebahq_comparison}(c). The more realistic synthesis images sampled from our model evidently indicate better generation quality.}
        \label{fig:qualitative_comparison_celebahq}
    \end{minipage}
    \hfill
    \begin{minipage}[t]{.48\linewidth}
        \vspace{0pt}
        \begin{minipage}[t]{.49\linewidth}
            \includegraphics[width=\linewidth]{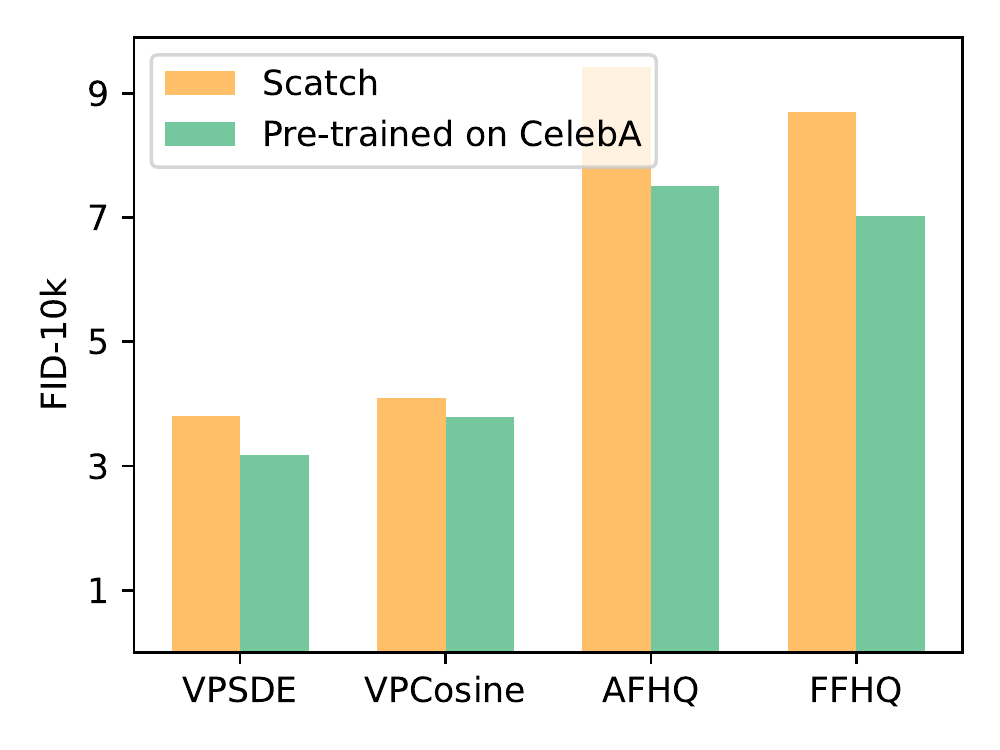}
            \subcaption{}
            
        \end{minipage}
        \hfill
        \begin{minipage}[t]{.49\linewidth}
            \includegraphics[width=\linewidth]{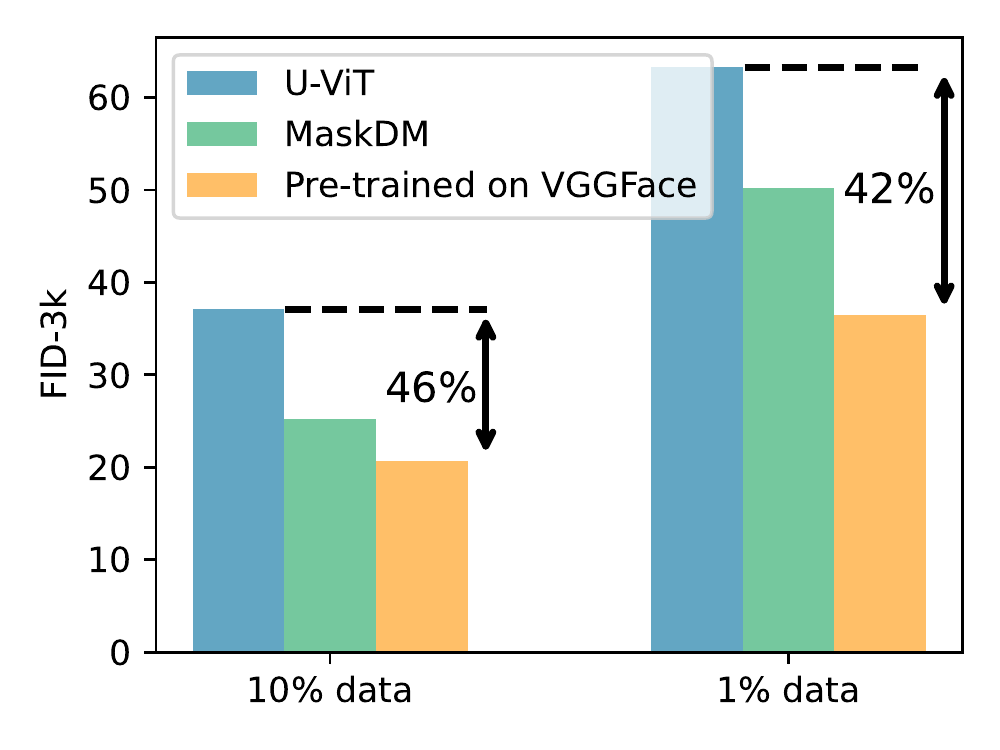}
            \subcaption{}
            
        \end{minipage}
        \caption{(a) Comparison between the performance of scratch-trained and adapted models. (b) Comparing models trained with no masking (yellow), masked pre-training (green), and cross-data fine-tuning (blue) on limited data.}
        \label{fig:generalizability}
        \label{fig:transfer_data}
        \label{fig:transfer_few_data}
    \end{minipage}
    
\end{figure*}

\subsection{Image Generation}
\label{sec:unconditional}

To thoroughly evaluate and demonstrates the efficacy of our proposed training paradigm, we conduct multiple experiments on image generation tasks with different resolutions, including the CelebA $64\times 64$ and $128\times 128$, LSUN Church $64\times 64$ and CelebA-HQ $256\times 256$. 

\textbf{Generation quality.} On the CelebA $64\times 64$ dataset, after loading the best pre-trained weight reported in Tab.\ref{tab:ablation_2}, we fine-tune MaskDM-S on 2 V100 GPUs and achieve an FID score of 2.2. The overall training takes approximately 2.09 V100 days. Our result significantly surpasses the FID scores reported by other works shown in Tab.\ref{tab:celeba_comparison}.

On the CelebA $128 \times 128$ dataset, to the best of our knowledge, GenViT~\cite{genvit} is the solitary ViT-based diffusion model built for this task.
We train a baseline U-ViT model with the identical settings (architecture, fine-tuning hyper-parameters and computational cost) as our MaskDM-B model and the objective detailed in Eq.~\ref{eq:lsimple} for comparison.
As previously mentioned in Sec.~\ref{sec:ablation}, we find that manually adjusting the mask rates and training steps during the entire pre-training stage leads to further model performance improvement. Specifically, we pre-train a MaskDM-B with 70\% mask rate and 30\% mask rate subsequently. Then we fine-tune the model and achieve the lowest FID score of 6.83 among all compared models. The training takes approximately 3.96 A100 days on 2 A100 GPUs, comprised of 3.26 A100 days pre-training and 0.69 A100 days fine-tuning.

On the LSUN Church $64\times 64$, we construct a MaskDM-S model with a 50\% mask rate and spent 1.63 V100 days for pre-training, followed by 0.17 V100 days of fine-tuning. Through a comparison between the trained MaskDM-S model and the baseline U-ViT model, we demonstrate a significant improvement of 23\% in the FID score. Furthermore, given the considerable difference between the LSUN Church dataset images and face images, this outcome convincingly showcases the generalizability of our proposed training methodology.

On the CelebA-HQ $256 \times 256$ dataset, we also manually adjust the mask rate and pre-train a MaskDM-B with 90\% mask rate and 50\% mask rate in a sequence. 
Thereafter, we fine-tune the model and achieve an FID score of 6.27. 
Training this MaskDM-B model takes 12.19 A100 days on 2 A100 GPUs, which comprises of 9.86 A100 days for pre-training and 2.33 A100 days for fine-tuning. 
We build a baseline model as previously done for CelebA $128\times128$, and only achieves an FID score of 24.83 at the training cost of 18.28 A100 days.
Comparing with other methods that either optimize models with adversarial training ~\citep{esser2021taming, karras2017progressive, xiao2021tackling} or training based on UNet\cite{ronneberger2015u} in raw pixel space~\citep{song2020score, kim2021soft, choi2022perception}, our MaskDM-B model achieves the best FID score utilizing U-ViT architecture with comparable model parameters, while maintaining a reasonable training cost.

\textbf{Comparison with LDMs.}
There are also studies~\citep{vahdat2021score, rombach2022high} on improving training efficiency for diffusion models by focusing on the latent space.
In comparison with these two methods,  our MaskDM-B model significantly outperforms LSGM~\cite{vahdat2021score}, but obtains worse performance than LDM~\cite{rombach2022high}. 
It is important to note that LDM~\cite{rombach2022high} utilizes an extra VAE~\cite{vae} as the feature extraction model, which is pre-trained on ImageNet $256 \times 256$ for hundreds of thousands of steps. In contrast, we only train one 102M-parameter ViT-based model consistently on CelebA-HQ $256 \times 256$ without any extra data. 
We anticipate our performance could be further enhanced by scaling up model parameters and incorporating advanced training techniques~\citep{EDM, ADM}. 

\textbf{Training efficiency.} We further present comparisons of the training efficiency between baseline and MaskDM models on CelebA and CelebA-HQ datasets in Fig.~\ref{fig:celebahq_comparison}.
For reaching similar FID scores (shown in vertical axes), MaskDM models can save 60\% ($\sim$5/12 in Fig.~\ref{fig:celebahq_comparison}(a)) to 80\% ($\sim$30/165 in Fig.~\ref{fig:celebahq_comparison}(c)) training hours than baseline models. 
Moreover, the reduction in training time increases as the data dimensionality increases.
We further select pairs of example human face images, as shown in Fig.~\ref{fig:qualitative_comparison_celebahq}, to provide a qualitative comparison between the synthesis results obtained by pre-trained and baseline model for reaching the same FID scores. 
The more realistic synthesis images sampled from MaskDM evidently indicate better generation quality.
We note that the pre-training curve is saturated for several steps. In practice, however, we find more pre-training time eventually leads to better FID scores, consistent with findings in Sec.\ref{sec:ablation}.

\subsection{Generalizability of MaskDM}
\label{sec:generalizability}

As previously mentioned in Section~\ref{sec:intro}, we expect a MaskDM model to have desirable generalizability to facilitate fine-tuning for various downstream image generation tasks.
In the following experiments, we assess the generalizability of MaskDM models by fine-tuning them with different datasets and training frameworks (Fig.~\ref{fig:transfer_data}(a)), and we also verify the benefit of pre-trained model for tasks confronted with data scarcity by fine-tuning with limited data (Fig.~\ref{fig:transfer_few_data}(b)). Training details are presented in Appendix~\ref{tab:appdix_hyper_of_fig_2_part1} and ~\ref{tab:appdix_hyper_of_fig_2_part2}.

We first choose CelebA $64\times 64$ as the source pre-training dataset and consider the scenario where the diffusion model training paradigms are not aligned across pre-training and fine-tuning (left bars in Fig.~\ref{fig:transfer_data}(a)). 
Specifically, we utilize DDPM in masked pre-training and adopt VPSDE~\cite{song2020score} or VPCosine~\cite{nichol2021improved} in fine-tuning.
We also collect the pre-trained models from Sec.\ref{sec:ablation} and fine-tune them on another two different datasets, i.e., FFHQ~\cite{karras2019style} and AFHQ~\cite{choi2020stargan}, with all images resized to the resolution of $64\times 64$.
This creates data distribution shifts between the pre-training and fine-tuning datasets. 
As shown in  Fig.~\ref{fig:transfer_data}(a)), when compared with models trained from scratch for each setting, pre-trained models demonstrate clear stronger generalizability for both training paradigm and data distribution shifts. 

Then we construct two small training datasets, one with 3000 images(10\%) and the other with 300 images(1\%), from the CelebA-HQ $256 \times 256$ dataset. 
For each dataset, we maintain similar computational expenses for training models from scratch and training MaskDM models.
As illustrated in Fig.~\ref{fig:transfer_few_data}(b)), the final FID scores evaluated on the complete CelebA-HQ $256 \times 256$ dataset demonstrate that our proposed training framework significantly improve the quality of generated images.

Moreover, we leverage another model, which is pre-trained on the VGGFace2 $256 \times 256$ dataset~\cite{cao2018vggface2} (containing approximately 3M training images) with a mask rate of 90\% for comparison. 
As shown in Fig.~\ref{fig:transfer_few_data}(b)), after loading the pre-trained weights, the FID score is substantially improved by 46\% and 42\% in experiments that contains 3000 and 300 images respectively. 
These improved results underscore the potential for utilizing a model which is sufficiently pre-trained for a generation task to tackle similar tasks that confront with data scarcity issues. 

\section{Related Work}

{\bf Efficient training for diffusion models.} There have been various research efforts in improving the training efficiency for diffusion models. 
One rapidly evolving research thread involves employing masking strategies during the training process, as exemplified by the masked autoencoder (MAE)~\cite{he2022masked} approach. 
Specifically, two recent studies~\citep{rombach2022high, vahdat2021score} propose latent diffusion models (LDM). These models apply the diffusion process on low-dimensional features extracted by a pre-trained VAE~\cite{vae}, reducing the cost of approximating extraneous details in the raw data. Based on LDM, subsequent studies propose MDT and MaskDiT, which incorporate masking into the latent space during training. 
However, the training complexity for building the feature extraction model (usually requires an extra large-scale training dataset such as ImageNet~\cite{deng2009imagenet} or OpenImage~\cite{OpenImages}) is often overlooked. 
Moreover, these methods often requires heavy computation costs for adapting to new tasks. 
In contrast, our method avoids dependencies on external models and maintains effectiveness even with limited data from downstream tasks. Also, our pre-trained MaskDM are considerably easier and efficient to generalize to various image generation tasks.

\textbf{ViT-based diffusion models.} Early research on diffusion models primarily utilized CNN architectures such as UNet. However, more recent studies~\citep{bao2022all, peebles2022scalable, cao2022exploring, genvit} have shown increasing interest in employing vision transformers (ViT) as the backbone for diffusion models. 
In comparison with CNN-based architecture, ViT-based architecture is readily scalable~\cite{peebles2022scalable} and compatible~\cite{bao2023one} with different data modalities. 
Notably, recent works~\citep{hoogeboom2023simple,zheng2023fast} have demonstrated the superiority of ViT-based diffusion models, such as MaskDiT, which achieved a FID score of 2.28 on ImageNet in 200 V100 days, while LDM (based on UNet) costed 271 V100 days to achieve a FID score of 3.60 on the same dataset. 
In a previous study~\cite{wang2023patch} parallel to our research, it was suggested to alternately train CNN-based diffusion models on both entire images and cropped patches. 
However, previous studies~\citep{bao2022all, peebles2022scalable, cao2022exploring, genvit} have analyzed the factors that affect the quality of generated samples and suggest smaller patch sizes correlate with better generated sample quality. 
This finding exacerbates ViT's computational drawbacks, including high CUDA memory usage and extended training times. 
In this work, since we aim at reducing the training costs spent on the cumbersome approximation of full-resolution data, our proposed training framework can effectively mitigate the computation challenge for training ViT-based diffusion models. 

\section{Conclusion}
In this work, a masked pre-training approach is proposed to improve the training efficiency for diffusion models in the context of image synthesis. 
We design a masked denoising score matching objective to guide the model for learning a primer distribution that shares some diverse and important features, conveyed in group of marginals, with the target data distribution. 
We empirically investigate various masking configurations for their impacts on model performance and training efficiency, and evaluate our approach using U-ViT for image synthesis in the pixel space on several different datasets. 
The evaluation results show that our approach substantially reduces the training cost while maintaining high generalization quality, outperforming the standard DDPM training method by a significant margin. 
We also conduct experiments for evaluating the generalizability of models pre-trained through our approach in the cases of training paradigm mismatch, data distribution shift, and limited training data. 
We demonstrate that our approach suffices to produce a pre-trained model with strong generalization capabilities. 
{
    \small
    \bibliographystyle{ieeenat_fullname}
    \bibliography{main}
}

\clearpage
\setcounter{page}{1}
\maketitlesupplementary

\section{Pre-training schedule}
\label{sec:pre-training converge}
\begin{figure}[h!]
    \centering
    \includegraphics[width=.7\linewidth]{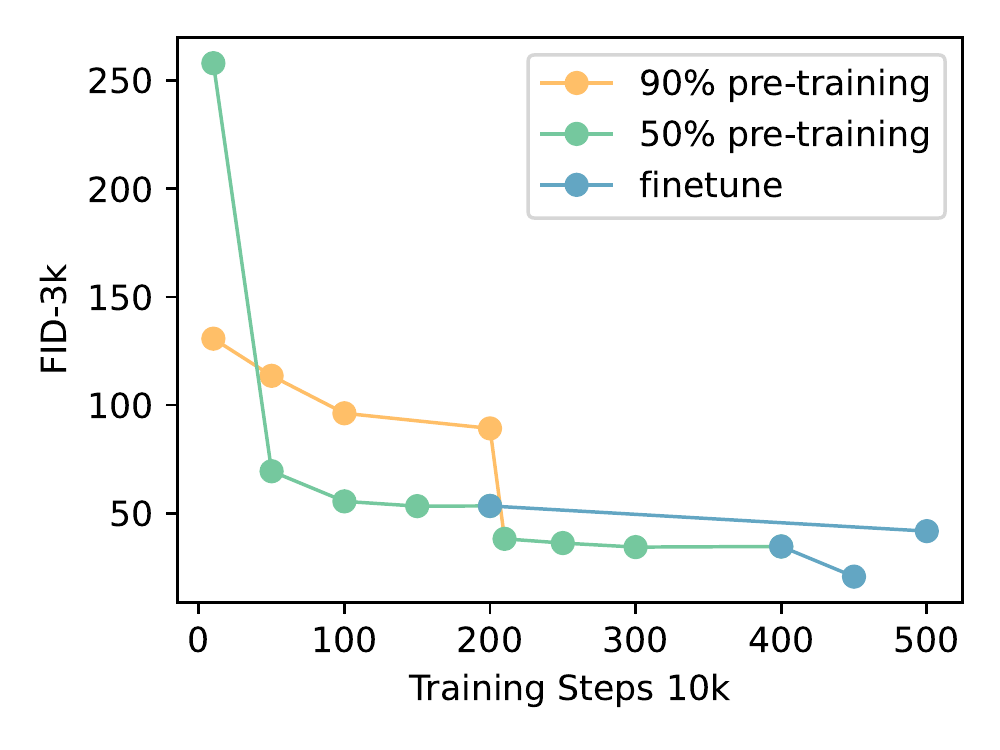}
    \caption{Comparison on different pre-training schedules.}
    \label{fig:appdix_pretraining_curve}
\end{figure}

As illustrated in Fig.\ref{fig:appdix_pretraining_curve}, during pre-training, the MaskDM model rapidly converges merely after 200k steps of training.

Besides, we demonstrate the performance of models that are pre-trained under two distinctive schedules: one is directly pre-trained at 50\% mask rate and the other is pre-trained at 90\% mask rate and subsequently at 50\% mask rate. We fine-tune both models using the same hyperparameters and maintain the overall training expense to be consistent across schedules. As a result, the model that is pre-trained progressively with two different mask rates achieves the best performance. More qualitative comparisons are presented in Fig.\ref{fig:appdix_block_50p}, Fig.\ref{fig:appdix_block_90p_50p_part1} and Fig.\ref{fig:appdix_block_90p_50p_part2}.

\section{The Pre-training instability at 90\% mask rate}

\label{section:appdix_stability}
\begin{figure*}
\centering
  \makebox[\textwidth]{
  \includegraphics[width=\textwidth]{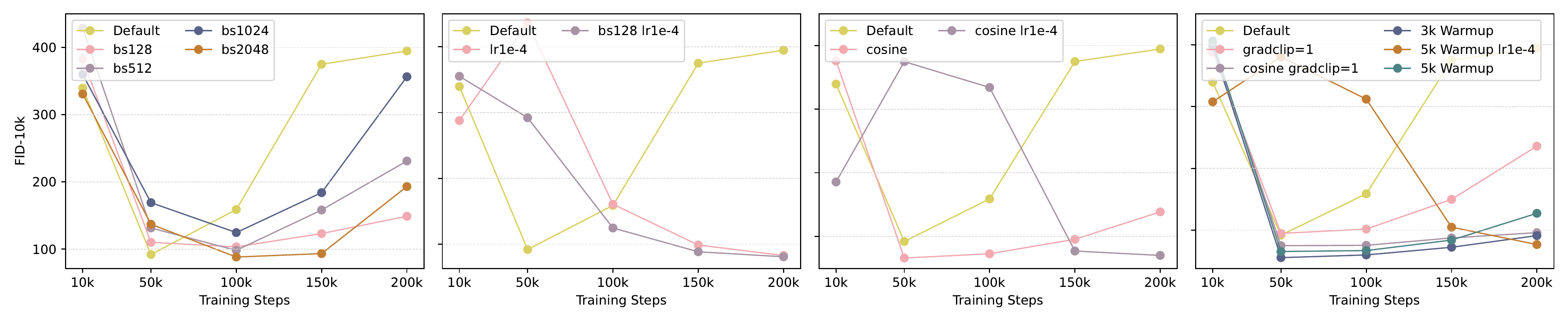}
  }
    \caption{Investigation into training instability. We maintain a fixed mask ratio of 90\% and adopt the parameters used in pre-training experiments at the 50\% mask ratio (See Tab.\ref{tab:appdix_hyper_of_table}) as default setting. The experiments are conducted on CelebA$64\times64$ dataset}
    \label{fig:instability}
\end{figure*}

In early experiments, we consistently observe that the model hardly converges when trained at 90\% mask ratio. Therefore, we investigate the impact of various factors on the pre-training process, including batch size, learning rate, noise schedule, gradient clipping, and learning rate schedule (Warmup). The default experiment setting is: lr=2e-4, batch size=256, linear noise schedule and no gradient clip or warmup.

As shown in Fig.\ref{fig:instability}, increasing the batch size from 128 to 2048 still results in divergence. It should be noted that the model demonstrate more stability when the batch size is set to 128, which is smaller than 256. A similar phenomenon is also observed in \cite{xiao2021tackling}.

Besides, when we reduce the learning rate from 2e-4 to 1e-4 and the batch size from 256 to 128, we observe a stable and gradual plateau in the FID score of the model after 200k training steps. However, this more conservative hyperparameter setting leads to a relatively slower convergence speed, necessitating a longer training time to achieve a similar level of performance for the model. 

Subsequently, we adopt a cosine schedule in place of the linear schedule, and the resulting FID score curve demonstrates the superiority of the cosine schedule in improving the training stability of the model. It effectively mitigates the convergence issues observed with the linear schedule. Additionally, we implement extra optimization strategies, such as Warmup and gradient clipping, which also contribute to a more stable convergence of the model.

\section{Model Configurations}
We present the configuration of our MaskDM models implemented in our experiments in Tab.\ref{tab:model_config}. Patch size is set to 4 in all experiments.
\begin{table}[h!]
    \centering
    \caption{Details of MaskDM models}
    \label{tab:model_config}
    \tablestyle{8pt}{1.05}
   \begin{tabular}{ c  c  c  c  c  c c}
        Model & Depth & Dim & MLP Dim & Heads & Params \\
        \toprule
        MaskDM-S  & 13  & 512 & 2048  & 8   & 44M   \\
        MaskDM-B  & 12  & 768 & 3172  & 12  & 102M  \\
        \bottomrule
    \end{tabular}
\end{table}

\section{Implementation Details}
In this section, we present training hyperparameters of our main experiments.
Specifically, in Tab.\ref{tab:appdix_hyper_of_table}, we present hyperparameters used in experiments in Tab.\ref{tab:ablation}, Tab.\ref{tab:celeba_comparison}, Tab.\ref{tab:celebahq128} and Tab.\ref{tab:celebahq_comparison}. In Tab.\ref{tab:appdix_hyper_of_fig_1}, we present hyperparameters used in Fig.\ref{fig:celebahq_comparison}. Finally, in Tab.\ref{tab:appdix_hyper_of_fig_2_part1} and Tab.\ref{tab:appdix_hyper_of_fig_2_part2}, we present hyperparameters used in Fig.\ref{fig:transfer_data} and Fig.\ref{fig:transfer_few_data} respectively.

\label{sec:implementation_details}
\begin{table*}
    \centering
    \begin{minipage}{\linewidth}
        \tablestyle{5pt}{1.2}
        \caption{hyperparameters for experiments in Tab.\ref{tab:ablation}, Tab.\ref{tab:celeba_comparison}, Tab.\ref{tab:celebahq128}, Tab.\ref{tab:celebahq_comparison}. Noticeably, on LSUN $64\times 64$, the baseline model is trained for 550k steps with identical hyperparameters that are used when fine-tuning the MaskDM model. }
        \label{tab:appdix_hyper_of_table}
        \begin{tabular*}{\linewidth}{ l  c c c  c  c  c  c c}
            Dataset & \multicolumn{4}{c}{CelebA$64\times 64$}  & CelebA$64\times 64$  & CelebA $128\times128$ & CelebA $256\times256$ & LSUN Church$64\times64$ \\
            \toprule
            Experiment      & \multicolumn{4}{c}{Tab.\ref{tab:ablation}} & Tab.\ref{tab:celeba_comparison} & Tab.\ref{tab:celebahq128} & Tab.\ref{tab:celebahq_comparison}  & Tab.\ref{tab:lsun_comparison}\\
            \midrule
            \multicolumn{9}{l}{\textbf{pre-train}} \\
            \midrule
            Masking         & \multicolumn{3}{c}{any}         & -               & 2x2 block-wise & 4x4 block-wise     & 4x4 block-wise  & 4x4 block-wise \\
            \midrule
            Mask rate       & 10\% & 50\%            & 90\%   & -               & 50\%           & 70\%, 50\%         & 90\%, 50\%      & 50\%  \\
            Lr              & 1e-4 & 2e-4            & 2e-4   & -               & 2e-4           & 2e-4, 1e-4         & 2e-4            & 2e-4  \\
            Batch size      & 128  & 256             & 512    & -               & 256            & 256, 128           & 128, 64         & 256   \\
            Steps           & 50k  & any             & 50k    & -               & 150k           & 50k, 350k          & 200k, 500k      & 500k \\
            Gradient clip   & \multicolumn{4}{c}{-}                             &  -             & 1.0, 1.0           & -, 1.0          & - \\
            Warmup          & \multicolumn{4}{c}{-}                             & 5k steps       & 5k steps, 5k steps & -, 5k steps     & 5k steps \\
            Noise schedule  & \multicolumn{3}{c}{Linear}      &                 & Linear         & Cosine             & Cosine          & Linear \\
            \midrule
            \multicolumn{9}{l}{\textbf{fine-tune}} \\
            \midrule
            Lr              & 1e-4    & 1e-4 & 1e-4  & 1e-4                     & 2e-4           & 5e-5             & 1e-5              & 2e-4\\
            Batch size      & 128     & 128  & 128   & 128                      & 128            & 64               & 32                & 128 \\
            Steps           & 200k    & 200k & 200k  & 250k                     & 350k           & 100k             & 100k              & 50k \\
            EMA setting     & \multicolumn{4}{c}{0.999 update every 1}          & 0.9999 update every 1   & 0.999 update every 1 & 0.999 update every 1 & 0.999 update every 1\\
            Gradient clip   & \multicolumn{4}{c}{-}                             &  -             & 1.0        & 1.0       & - \\
            Warmup          & \multicolumn{4}{c}{-}                             & 5k steps       & 5k steps   & 5k steps  & 5k steps \\
            \midrule
            \multicolumn{9}{l}{\textbf{shared parameters}} \\
            \midrule
            Model           & \multicolumn{4}{c}{MaskDM-S}                       & MaskDM-S        & MaskDM-B & MaskDM-B & MaskDM-S\\
            Noise schedule  & \multicolumn{4}{c}{Linear}                         & VPSDE           & Cosine   & Cosine & Linear\\
            Horizontal flip & \multicolumn{4}{c}{-}                              & 0.5             & -        & -      & - \\
            \midrule
            \multicolumn{9}{l}{\textbf{sampling}} \\
            \midrule
            Sampler         & \multicolumn{4}{c}{DDIM}                          & Euler-Maruyama & DDIM      & DDIM & DDIM \\
            Sampling steps  & \multicolumn{4}{c}{500 steps}                     & 1000 steps     & 500 steps & 500 steps & 500 steps \\
            Num of samples  & \multicolumn{4}{c}{10k}                           & 50k            & 50k       & 50k   & 50k\\
            \bottomrule
        \end{tabular*}
    \end{minipage}
\end{table*}

\begin{table*}
    \centering
    \begin{minipage}{.8\linewidth}
        \centering
        \caption{hyperparameters for MaskDM and baseline models in Fig.\ref{fig:celebahq_comparison} and baseline models in Tab.\ref{tab:celeba_comparison}, Tab.\ref{tab:celebahq128}, Tab.\ref{tab:celebahq_comparison}. For the baseline model, we use exactly the same hyperparameter settings as in the fine-tuning of MaskDM. 
        Additionally, the training step is set to 250k, 550k and 800k steps for baseline model on CelebA$64\times 64$, CelebA $128\times 128$, CelebA-HQ$256\times 256$, respectively, maintaining a consistent computational cost with the MaskDM counterpart.
        }
        \label{tab:appdix_hyper_of_fig_1}
        \begin{tabular}{ l  c c c}
            Dataset         & CelebA$64\times 64$    & CelebA $128\times128$ & CelebA $256\times256$  \\
            \toprule
            \multicolumn{4}{l}{\textbf{pre-train}} \\
            \midrule
            Masking         & 2x2 block-wise & 4x4 block-wise        & 4x4 block-wise  \\
            \midrule
            Mask rate       & 50\%           & 70\%, 50\%            & 90\%, 50\%     \\
            Lr              & 2e-4           & 2e-4, 1e-4            & 2e-4           \\
            Batch size      & 256            & 256, 128              & 128, 64        \\
            Steps           & 50k            & 50k, 200k             & 200k, 500k     \\
            Gradient clip   & -              & 1.0, 1.0              & -, 1.0         \\
            Warmup          & -              & 5k steps, 5k steps    & -, 5k steps    \\
            \midrule
            \multicolumn{4}{l}{\textbf{fine-tune}} \\
            \midrule
            Lr              & 1e-4           & 5e-5             & 1e-5  \\
            Batch size      & 128            & 64               & 32    \\
            Steps           & 200k           & 100k             & 100k  \\
            EMA setting     & 0.9999 update every 1   & 0.999 update every 1 & 0.999 update every 1\\
            Gradient clip   & -              & 1.0              & 1.0        \\
            Warmup          & -              & 5k steps         & 5k steps\\
            \midrule
            \multicolumn{4}{l}{\textbf{shared parameters}} \\
            \midrule
            Model           & MaskDM-S       & MaskDM-B & MaskDM-B\\
            Noise schedule  & Linear         & Cosine   & Cosine\\
            Horizontal flip & -              & -        & -   \\
            \midrule
            \multicolumn{4}{l}{\textbf{sampling}} \\
            \midrule
            Sampler          & DDIM          & DDIM      & DDIM \\
            Sampling steps   & 500 steps     & 250 steps & 250 steps \\
            Num of samples   & 10k           & 10k       & 3k  \\
            \bottomrule
        \end{tabular}
    \end{minipage}
\end{table*}

\begin{table*}
    \centering
    \begin{minipage}{.8\linewidth}
        \centering
        \caption{hyperparameters of experiments in Fig.\ref{fig:transfer_data}. Following the parameters used during fine-tuning MaskDM, the baseline models are trained for 250k steps. We employ DPM-Solver~\cite{lu2022dpm} to generate samples on CelebA$64\times 64$ dataset.}
        \label{tab:appdix_hyper_of_fig_2_part1}
        \begin{tabular}{ l  c c c }
            Dataset         & CelebA$64\times 64$   & FFHQ $64\times 64$ & AFHQ $64\times64$ \\
            \toprule
            \multicolumn{4}{l}{\textbf{pre-train}} \\
            \midrule
            Masking         & \multicolumn{3}{c}{2x2 block-wise}  \\
            \midrule
            Mask rate       & \multicolumn{3}{c}{50\%}    \\
            Lr              & \multicolumn{3}{c}{2e-4}    \\
            Batch size      & \multicolumn{3}{c}{256}     \\
            Steps           & \multicolumn{3}{c}{50k}     \\
            Noise schedule  & \multicolumn{3}{c}{Linear}  \\ 
            \midrule
            \multicolumn{4}{l}{\textbf{fine-tune}} \\
            \midrule
            Lr              & \multicolumn{3}{c}{1e-4}    \\
            Batch size      & \multicolumn{3}{c}{128}     \\
            Steps           & \multicolumn{3}{c}{200k}     \\
            EMA setting     & \multicolumn{3}{c}{0.999 update every} \\
            \midrule
            \multicolumn{4}{l}{\textbf{shared parameters}} \\
            \midrule
            Model           & \multicolumn{3}{c}{MaskDM-S} \\
            Noise schedule  & VPSDE, VPCosine & Linear & Linear  \\ 
            Horizontal flip & 0.5              & -        & -   \\
            \midrule
            \multicolumn{4}{l}{\textbf{sampling}} \\
            \midrule
            Sampler          & DPM-Solver & DDIM      & DDIM \\
            Sampling steps   & 50 steps          & 500 steps & 500 steps \\
            Num of samples   & 10k               & 10k       & 10k  \\
            \bottomrule
        \end{tabular}
    \end{minipage}
\end{table*}

\begin{table*}
    \centering
    \begin{minipage}{.8\linewidth}
        \centering
        \caption{hyperparameters of experiments in Fig.\ref{fig:transfer_few_data}. Following the parameters used when fine-tuning MaskDM, the baseline models are trained for 200k steps.}
        \label{tab:appdix_hyper_of_fig_2_part2}
        \begin{tabular}{ l  c c }
            Dataset         & CelebA-HQ $256\times256$ 10\% or 1\%    & VGGFace2 $256\times 256$ \\
            \toprule
            \multicolumn{3}{l}{\textbf{pre-train}} \\
            \midrule
            Masking         &   4x4 block-wise   &   4x4 block-wise  \\
            \midrule
            Mask rate       &   50\%             &   90\% \\
            Lr              &   2e-4             &   2e-4  \\
            Batch size      &   64               &   256  \\
            Steps           &   200k             &   200k  \\
            Noise schedule  &   Cosine           &   Cosine   \\ 
            \midrule
            \multicolumn{3}{l}{\textbf{fine-tune}} \\
            \midrule
            Lr              & \multicolumn{2}{c}{5e-4} \\
            Batch size      & \multicolumn{2}{c}{64}   \\
            Steps           & \multicolumn{2}{c}{50k}   \\
            EMA setting     & \multicolumn{2}{c}{0.999 update every 1} \\
            \midrule
            \multicolumn{3}{l}{\textbf{shared parameters}} \\
            \midrule
            Model           & \multicolumn{2}{c}{MaskDM-S} \\
            Noise schedule  & \multicolumn{2}{c}{Cosine}  \\ 
            Horizontal flip & \multicolumn{2}{c}{0.5}     \\
            \midrule
            \multicolumn{3}{l}{\textbf{sampling}} \\
            \midrule
            Sampler          & \multicolumn{2}{c}{DDIM}\\
            Sampling steps   & \multicolumn{2}{c}{250 steps} \\
            Num of samples   & \multicolumn{2}{c}{3k}  \\
            \bottomrule
        \end{tabular}
    \end{minipage}
\end{table*}

\textbf{Lower LR used in 10\% masked pre-training in Tab.\ref{tab:ablation}.} In early experiments, we observe that the model yields a poor performance when the learning rate is set to 2e-4, using 128 batch size. Therefore, we scale the learning rate linearly according to the batch size and use 1e-4 in our experiments.

\section{Additional Visualization Results}
\begin{figure*}[h!]
    \centering
    \includegraphics[width=\linewidth]{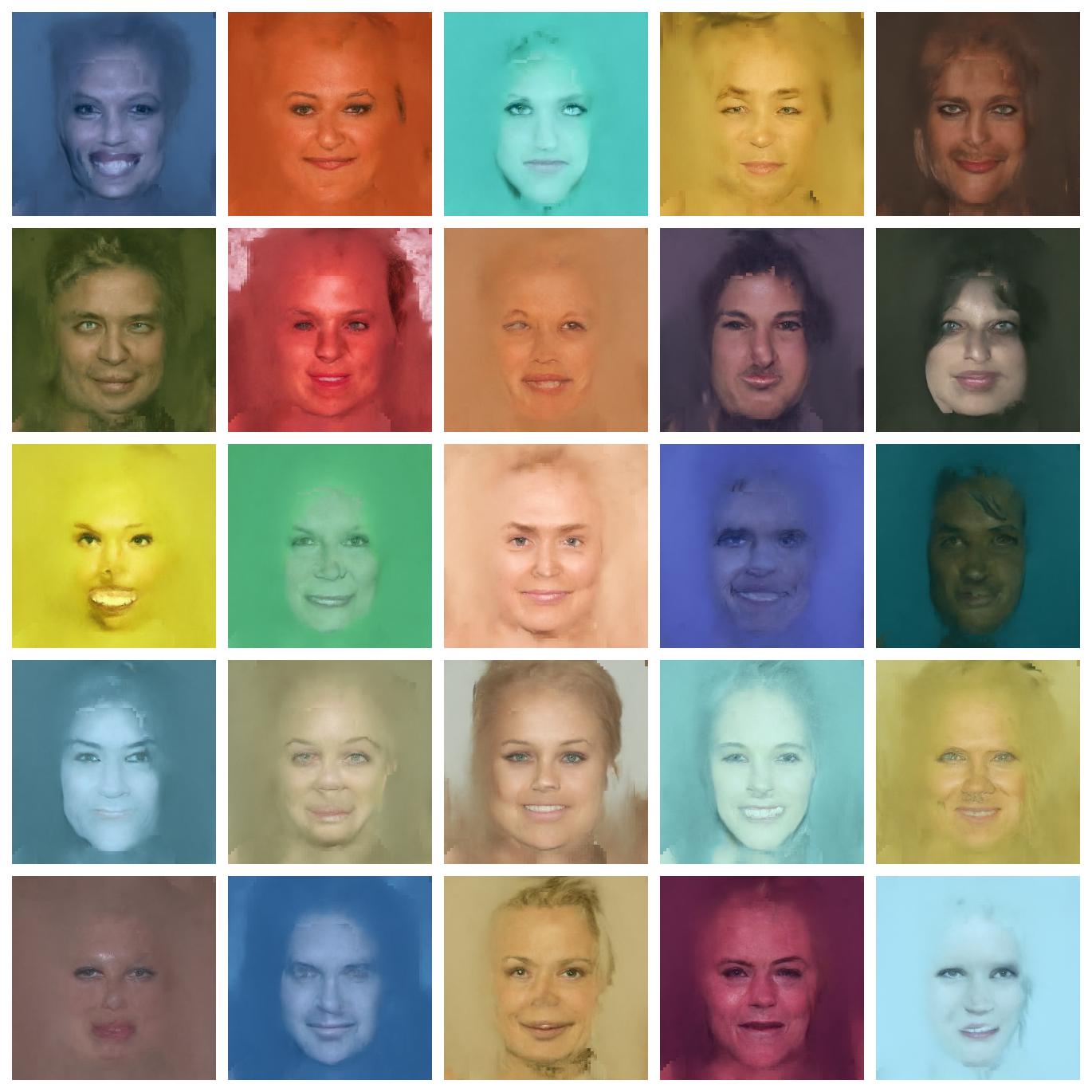}
    \caption{Samples generated by a MaskDM model pre-trained on CelebA-HQ$256\times 256$ given a masking strategy of cropping and 90\% mask rate.}
    \label{fig:appdix_cropping}
\end{figure*}

\begin{figure*}[h!]
    \centering
    \includegraphics[width=\linewidth]{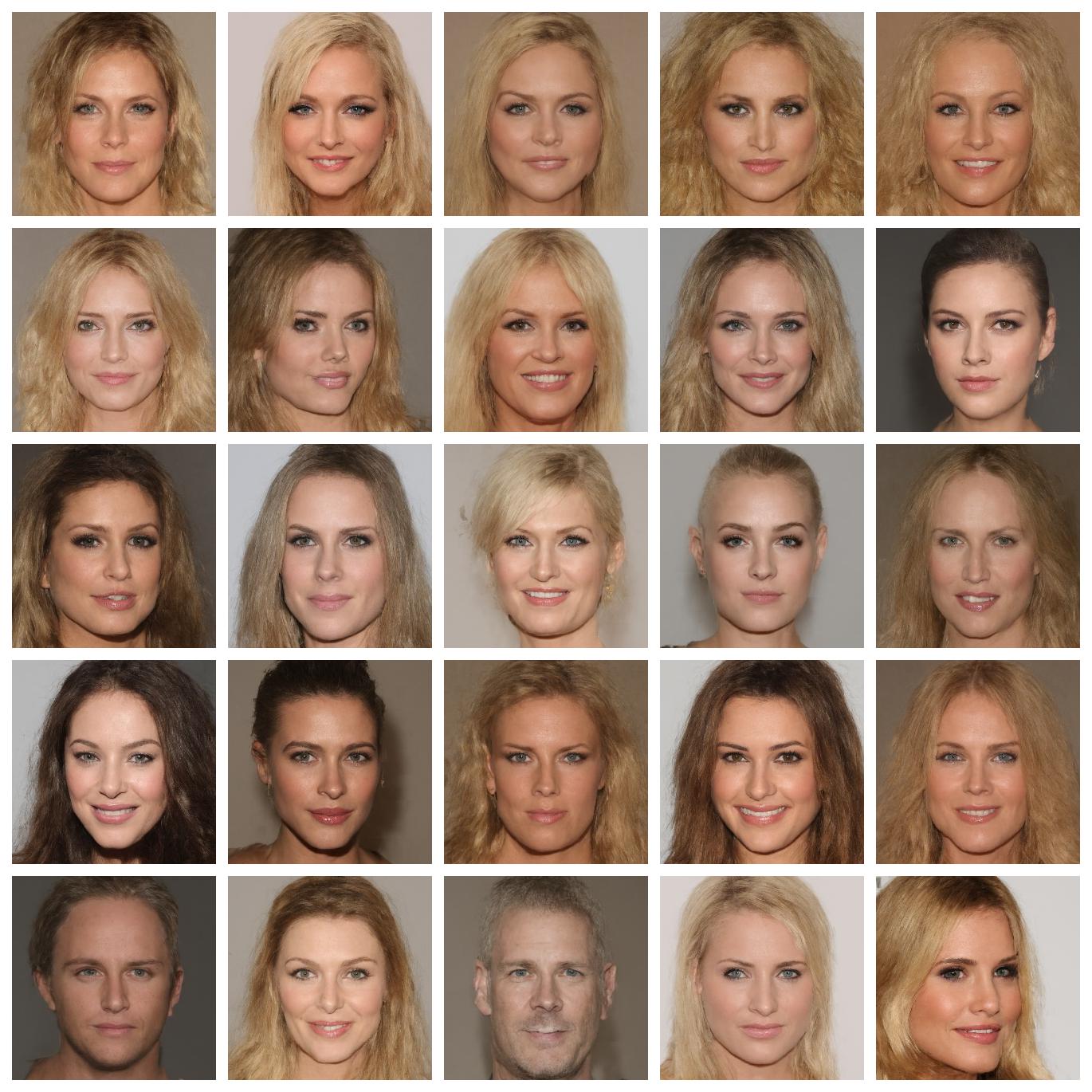}
    \caption{Uncurated samples generated by a MaskDM model pre-trained on CelebA-HQ$256\times 256$ (4x4 block-wise masking and 90\% mask rate). }
    \label{fig:appdix_block_90p}
\end{figure*}

\begin{figure*}[h!]
    \centering
    \includegraphics[width=\linewidth]{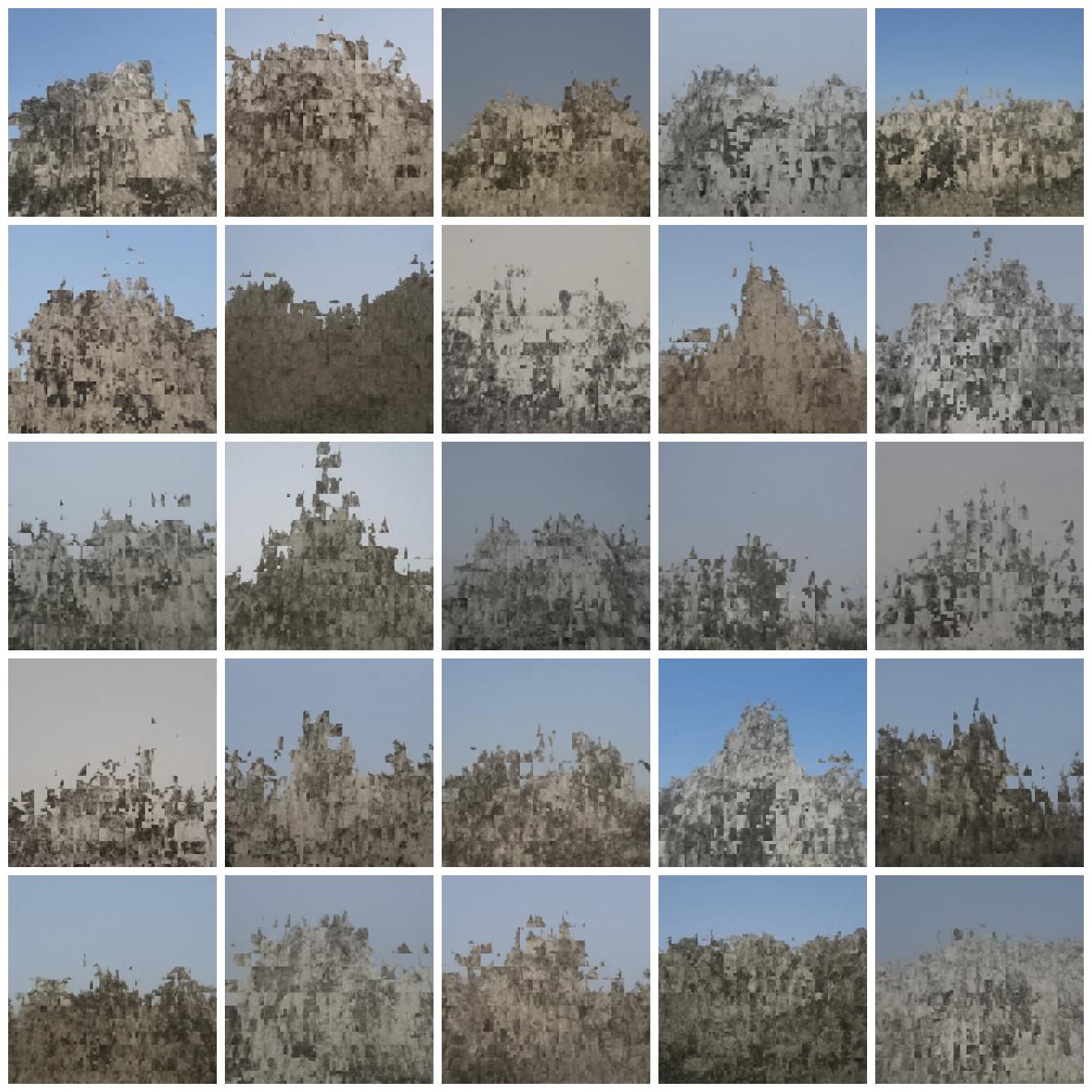}
    \caption{Uncurated samples generated by a MaskDM model pre-trained on LSUN Church$256\times 256$ (4x4 block-wise masking and 90\% mask rate). }
    \label{fig:appdix_block_90p_church}
\end{figure*}

\begin{figure*}[h!]
    \centering
    \includegraphics[width=\linewidth]{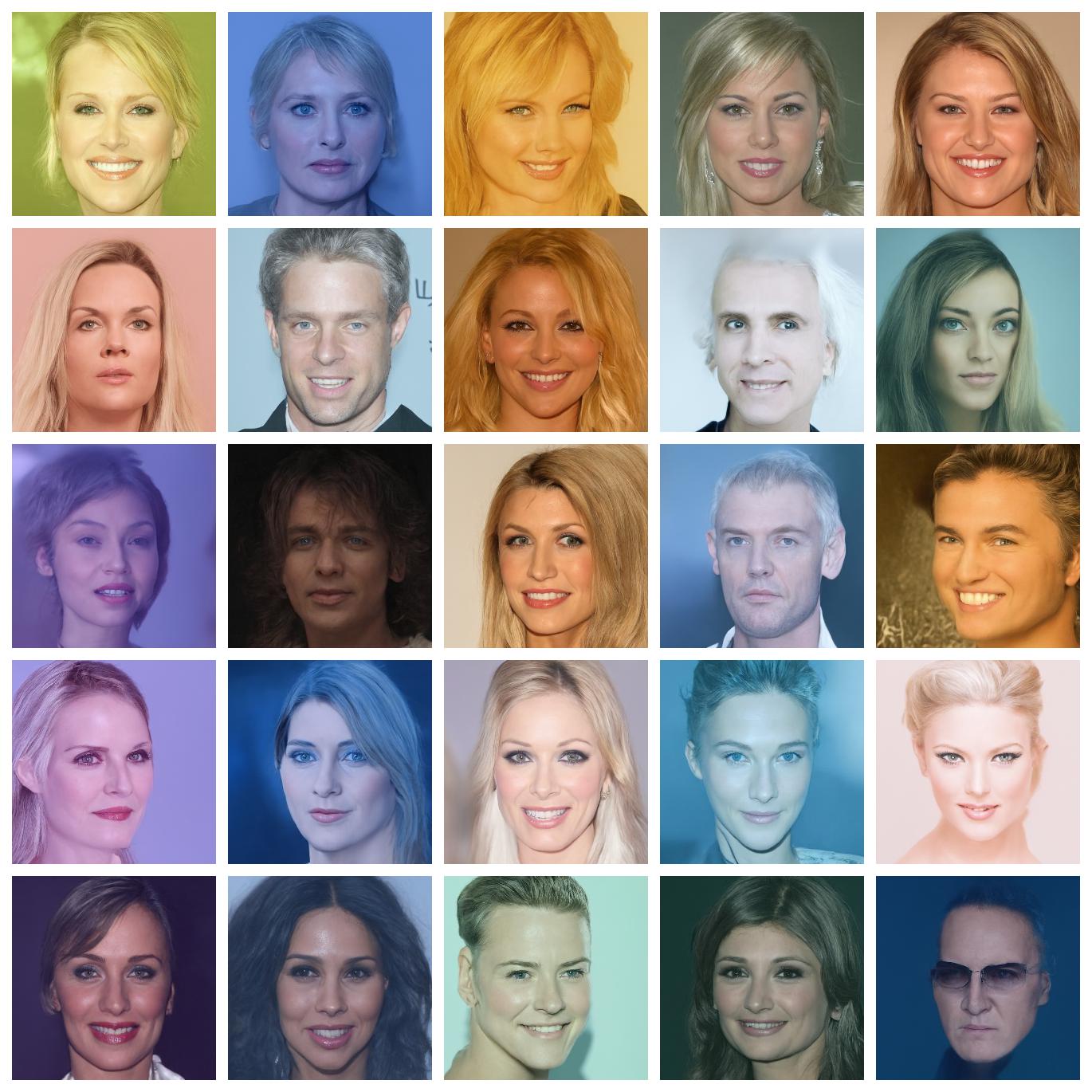}
    \caption{Uncurated samples generated by a MaskDM model pre-trained on CelebA-HQ$256\times 256$ (4x4 block-wise masking and 50\% mask rate). }
    \label{fig:appdix_block_50p}
\end{figure*}

\begin{figure*}[h!]
    \centering
    \includegraphics[width=\linewidth]{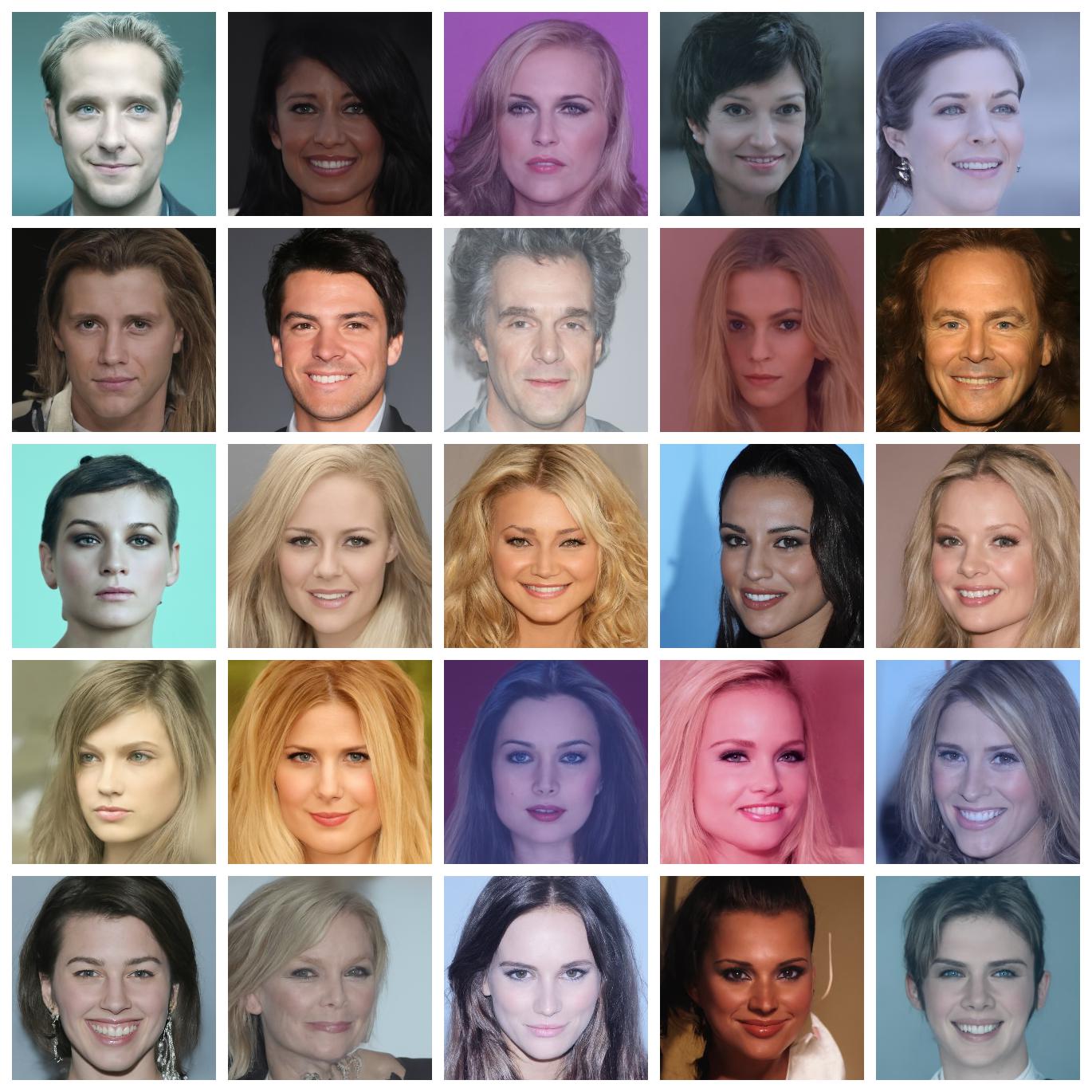}
    \caption{Uncurated samples generated by a MaskDM model pre-trained on CelebA-HQ$256\times 256$, given the configuration of 4x4 block-wise masking and 50\% mask rate, after loading weights that are pre-trained at 90\% mask rate. The pre-training at 50\% mask rate takes 100k steps.}
    \label{fig:appdix_block_90p_50p_part1}
\end{figure*}

\begin{figure*}[h!]
    \centering
    \includegraphics[width=\linewidth]{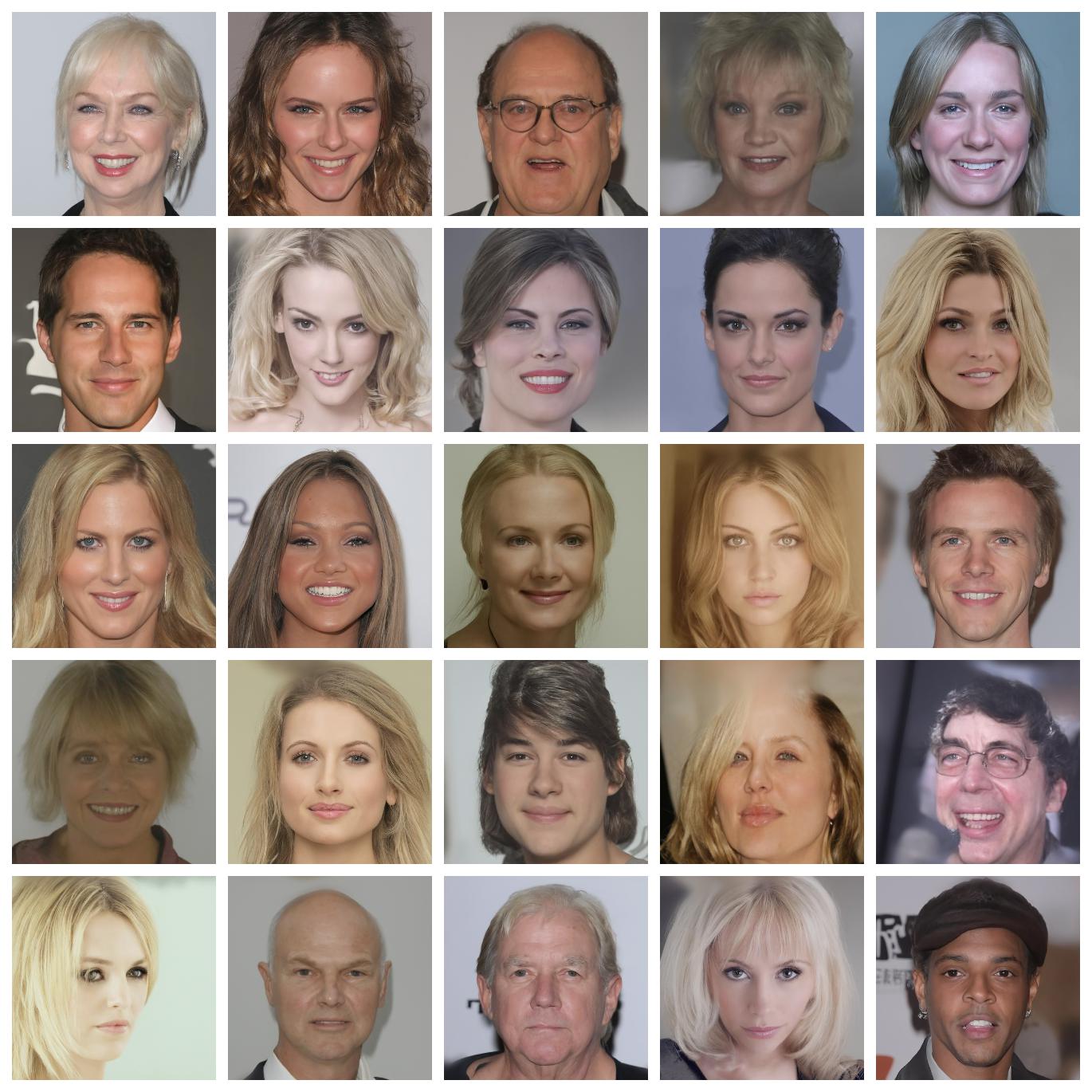}
    \caption{ Same setting as Fig.\ref{fig:appdix_block_90p_50p}. The pre-training at 50\% mask rate takes 500k steps.}
    \label{fig:appdix_block_90p_50p_part2}
\end{figure*}

\begin{figure*}[h!]
    \centering
    \includegraphics[width=\linewidth]{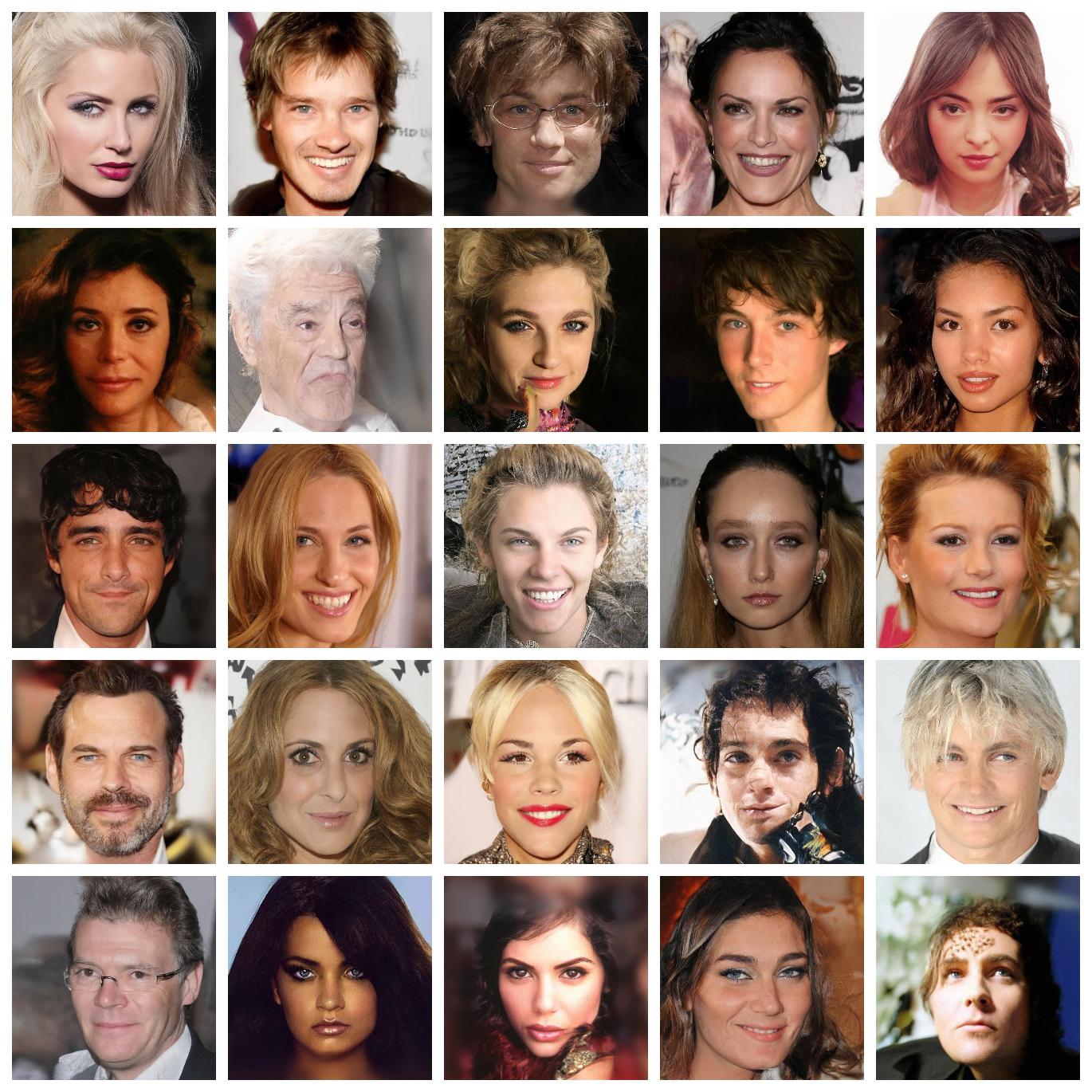}
    \caption{Uncurated samples generated by our MaskDM-B model in Tab.\ref{tab:celebahq_comparison}.}
    \label{fig:appdix_celebahq}
\end{figure*}

\begin{figure*}
    \begin{subfigure}{\linewidth}
        \begin{subfigure}{.5\linewidth}
         \centering
        \includegraphics[width=\linewidth]{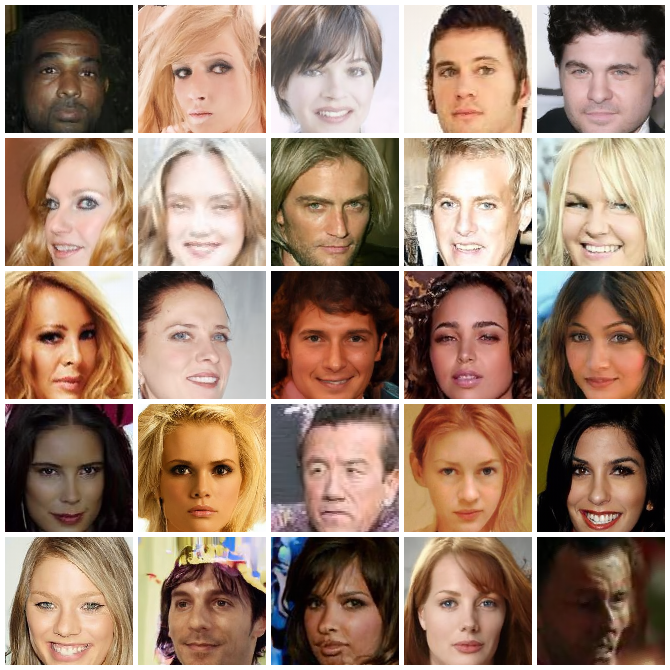}
        \end{subfigure}%
        \begin{subfigure}{.5\linewidth}
             \centering
            \includegraphics[width=\linewidth]{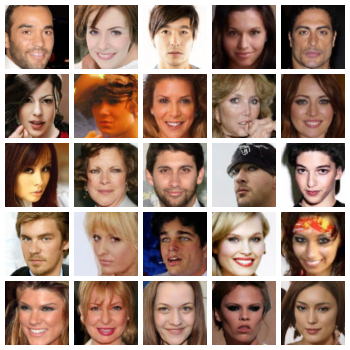}
        \end{subfigure}%
    \end{subfigure}
    \\
    \begin{subfigure}{\linewidth}
        \begin{subfigure}{.5\linewidth}
        \centering
        \includegraphics[width=\linewidth]{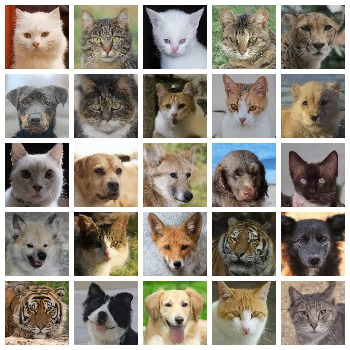}
        \end{subfigure}%
        \begin{subfigure}{.5\linewidth}
            \centering
            \includegraphics[width=\linewidth]{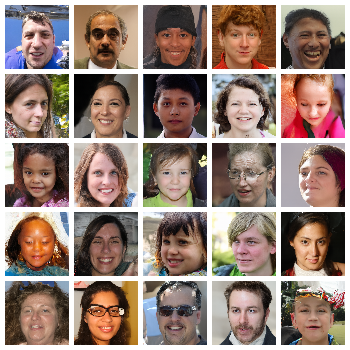}
        \end{subfigure}    
    \end{subfigure}

    \caption{Uncurated samples generated by our MaskDM-S models.}
    \label{fig:appdix_misc}
\end{figure*}

We further present more qualitative results in this section.
We display samples generated by MaskDM model that is: pre-trained with 90\% cropping on CelebA-HQ $256\times 256$ in Fig.\ref{fig:appdix_cropping},
pre-trained with 90\% 4x4 block-wise masking on CelebA-HQ $256\times 256$ in Fig.\ref{fig:appdix_block_90p},
pre-trained with 90\% 4x4 block-wise masking on LSUN Church $256\times 256$ in Fig.\ref{fig:appdix_block_90p_church},
pre-trained with 50\% 4x4 block-wise masking on CelebA-HQ $256\times 256$ in Fig.\ref{fig:appdix_block_50p},
pre-trained subsequently with 90\% and 50\% 4x4 block-wise masking on CelebA-HQ $256\times 256$ in Fig.\ref{fig:appdix_block_90p_50p_part1} and Fig.\ref{fig:appdix_block_90p_50p_part2},
fine-tuned on CelebA-HQ $256\times256$ in Fig.\ref{fig:appdix_celebahq},
and fine-tuned on other datasets in Fig.\ref{fig:appdix_misc}.

\end{document}